\documentclass[1p]{elsarticle}  

\usepackage{hyperref}
\hypersetup{
   breaklinks=true,   
   colorlinks=true,   
   pdfusetitle=true   
}

\usepackage{amsmath,amssymb}
\makeatletter
\let\MYcaption\@makecaption
\makeatother

\usepackage{subcaption}
\makeatletter
\let\@makecaption\MYcaption
\makeatother
\usepackage{multirow}

\usepackage{color}
\usepackage{ulem}
\newcommand{\Erasesue}{\bgroup\markoverwith{\textcolor{green}{\rule[.5ex]{2pt}{0.4pt}}}\ULon}
\newcommand{\Erasesaw}{\bgroup\markoverwith{\textcolor{blue}{\rule[.5ex]{2pt}{0.4pt}}}\ULon}
\newcommand{\Eraseike}{\bgroup\markoverwith{\textcolor{magenta}{\rule[.5ex]{2pt}{0.4pt}}}\ULon}
\newcommand{\Add}[1]{\textcolor{black}{#1}}
\newcommand{\Erase}[1]{\if0{#1}\fi}

\biboptions{authoryear}

\begin{document}

\begin{frontmatter}

\title{S$^3$NN: Time Step Reduction of Spiking Surrogate Gradients for Training Energy Efficient Single-Step Spiking Neural Networks\tnoteref{mytitlenote}}

\author[address1]{Kazuma Suetake}
\author[address2]{Shin-ichi Ikegawa}
\author[address1]{Ryuji Saiin}
\author[address2]{Yoshihide Sawada\corref{mycorrespondingauthor}}
\ead{yoshihide.sawada@aisin.co.jp}

\cortext[mycorrespondingauthor]{Corresponding author}
\address[address1]{AISIN SOFTWARE, Aichi, Japan}
\address[address2]{Tokyo Research Center, AISIN, Tokyo, Japan}


\begin{abstract}
As the scales of neural networks increase, techniques that enable them to run with low computational cost and energy efficiency are required.
From such demands, various efficient neural network paradigms, such as spiking neural networks (SNNs) or binary neural networks (BNNs), have been proposed. However, they have sticky drawbacks, such as degraded inference accuracy and latency.
To solve these problems, we propose a single-step spiking neural network (S$^3$NN), an energy-efficient neural network with low computational cost and high precision.
The proposed S$^3$NN processes the information between hidden layers by spikes as SNNs.
Nevertheless, it has no temporal dimension so that there is no latency within training and inference phases as BNNs.
Thus, the proposed S$^3$NN has a lower computational cost than SNNs that require time-series processing.
However, S$^3$NN cannot adopt na\"{i}ve backpropagation algorithms due to the non-differentiability nature of spikes.
We deduce a suitable neuron model by reducing the surrogate gradient for multi-time step SNNs to a single-time step.
We experimentally demonstrated that the obtained surrogate gradient allows S$^3$NN to be trained appropriately.
We also showed that the proposed S$^3$NN could achieve comparable accuracy to full-precision networks while being highly energy-efficient.
\end{abstract}

\begin{keyword}
Spiking Neural Network\sep Binary Neural Network\sep Surrogate Gradient\sep Energy Efficiency\sep Single-Time Step.
\end{keyword}

\end{frontmatter}


\section{Introduction}
\label{sec:introduction}
As neural networks become much deeper, there is a demand for technology that can drive them with high accuracy using fewer computing resources.
Additionally, there is a need for stronger energy-efficient neural networks than ever, as it is essential to reduce CO$_2$ emissions.
One of such solutions is spiking neural networks (SNNs).
SNNs are more bio-plausible neural networks designed to run energy efficiently on neuromorphic chips~\citep{akopyan2015truenorth,davies2018loihi} or field-programmable gate array (FPGA)~\citep{maguire2007challenges,misra2010artificial} by asynchronous processing.
Other solutions include binary neural networks (BNNs) that binarize parameters without significantly changing the network architecture. However, they have some disadvantages ({\it e.g.}, computational efficiency and inference accuracy) compared to standard artificial neural networks (ANNs). {\it Our goal is to develop a neural network with low computational cost, energy efficiency, and high precision.}

SNNs propagate the information as a spike ({\it i.e.}, $\{0, 1\}$), with the weights remaining real-valued.
Due to this concept, SNNs can replace the multiply-accumulate (MAC) operations with additive operations and are suitable for asynchronous processing.
However, SNNs are designed for the time-series processing of spikes ($\{0, 1\}^T$, where $T$ is the number of time steps).
This design makes the inference and training time longer than that of ANNs. 

Unlike SNNs, BNNs propagate the information as a binary ({\it i.e.}, $\{\pm 1\}$) by binarizing weights and inputs/outputs of each layer.
Additionally, algorithms that excel at binary calculations ({\it e.g.}, xnor or popcount) improve computational efficiency.
Meanwhile, these binarization techniques reduce the inference accuracy. 

Although SNNs and BNNs have different properties, they use similar neuron models that are difficult to train.
Their neurons are non-differential and cannot use backpropagation directly because they use binary or spike information.
Therefore, several surrogate gradients have been proposed to overcome this limitation.
A surrogate gradient is an approach where backpropagation is made possible by approximating the gradient of the activation function.
In particular, SNNs use more complex temporal-dependent nonlinear spiking neuron models; thus, highly technical surrogate gradients have been proposed.
However, several previous studies have conducted various experiments for static tasks that do not require time-series processing~\citep{zhang2020temporal,zheng2021going,chowdhury2021one}. This unnecessarily complicated procedure leads to high computational costs.

In this study, we consider {\it SNNs with a limit of time steps $\to 1$} to obtain a neural network for static tasks that excels in computational and energy efficiencies and accuracy.
This limiting trick eliminates the need for multi-step training/inference of typical SNNs and significantly reduces the computational cost.
Furthermore, we show that limiting time steps $\to 1$ for a spiking neuron model returns it to a step activation function (Heaviside function).
Therefore, we can obtain the neural networks of roughly the same inference cost as BNNs.
The difference is that our model's weights maintain real-valued.
Although our model cannot use binary calculations, it is still computationally efficient because of replacing MAC operations with additive operations.
It is also promising in terms of energy efficiency and inference accuracy (Table~\ref{tbl:comparison-overview}).

\begin{table}[t]
\centering
\caption{Comparison overview of the proposed S$^3$NN with existing general BNNs and SNNs.
Our model is superior to these in terms of computational efficiency, energy efficiency ({\it{i.e.}}, whether an energy-efficient asynchronous processing is suitable or not), and accuracy.}
\begin{tabular}{lccc}
\hline
& BNNs & SNNs & S$^3$NN\\
\hline
Input/Output & $\{\pm 1\}$ & $\{0,1\}^T$ & $\{0,1\}$ \rule[0mm]{0mm}{4mm} \\
Weights & $\{\pm 1\}$ & $\boldsymbol{R}$ & $\boldsymbol{R}$ \\
\hline
Computational efficiency & $\checkmark$ & $\times$ & $\checkmark$ \\
Energy efficiency & $\times$ & $\checkmark$ & $\checkmark$ \\
Inference accuracy & $\times$ & $\checkmark$ & $\checkmark$ \\
\hline
\end{tabular}
\label{tbl:comparison-overview}
\end{table}

\begin{figure}[t]
\begin{minipage}[b]{0.3\linewidth}
    \centering
    \includegraphics[keepaspectratio,width=\linewidth]{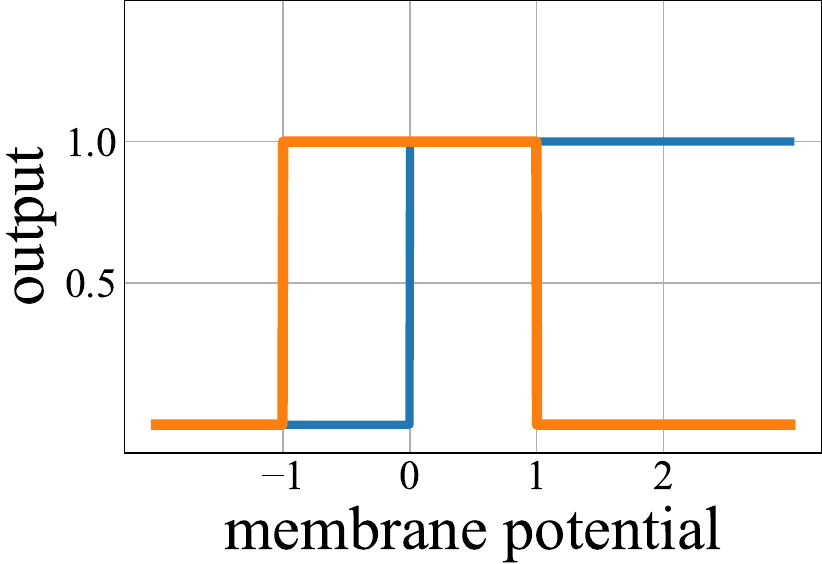}\\
    \subcaption{STE-b}
    \label{fig:surrogate-gradients-ste-b}
\end{minipage}
\hfill
\begin{minipage}[b]{0.3\linewidth}
    \centering
    \includegraphics[keepaspectratio,width=\linewidth]{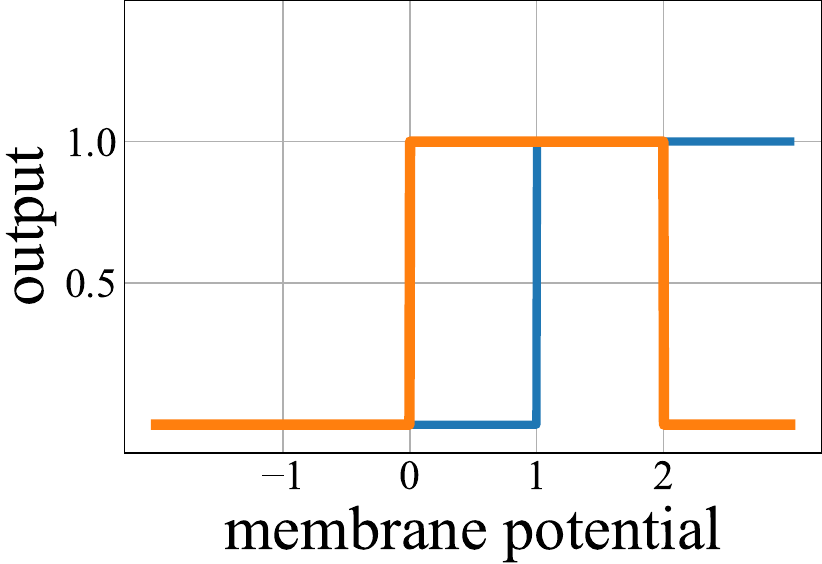}\\
    \subcaption{STE-s}
    \label{fig:surrogate-gradients-ste-s}
\end{minipage}
\hfill
\begin{minipage}[b]{0.3\linewidth}
    \centering
    \includegraphics[keepaspectratio,width=\linewidth]{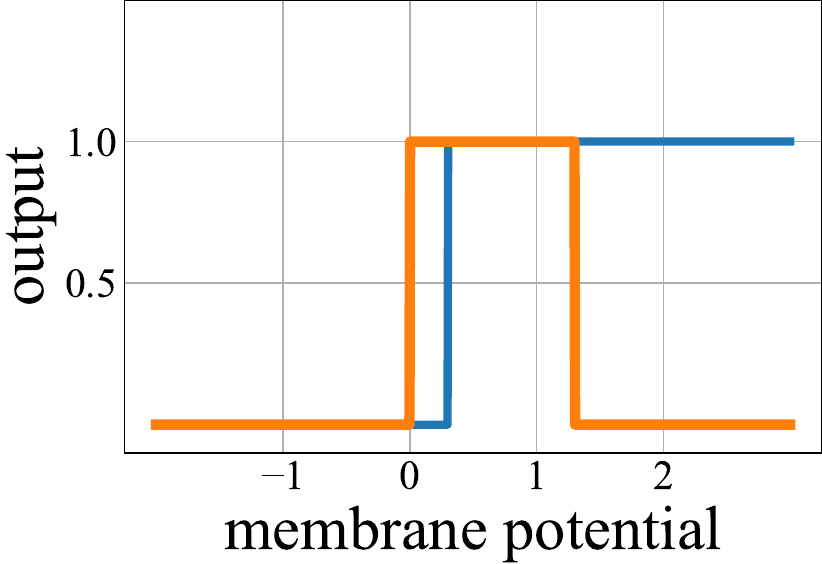}\\
    \subcaption{SiBNN}
    \label{fig:surrogate-gradients-sibnn}
\end{minipage}
\\
\begin{minipage}[b]{0.3\linewidth}
    \centering
    \includegraphics[keepaspectratio,width=\linewidth]{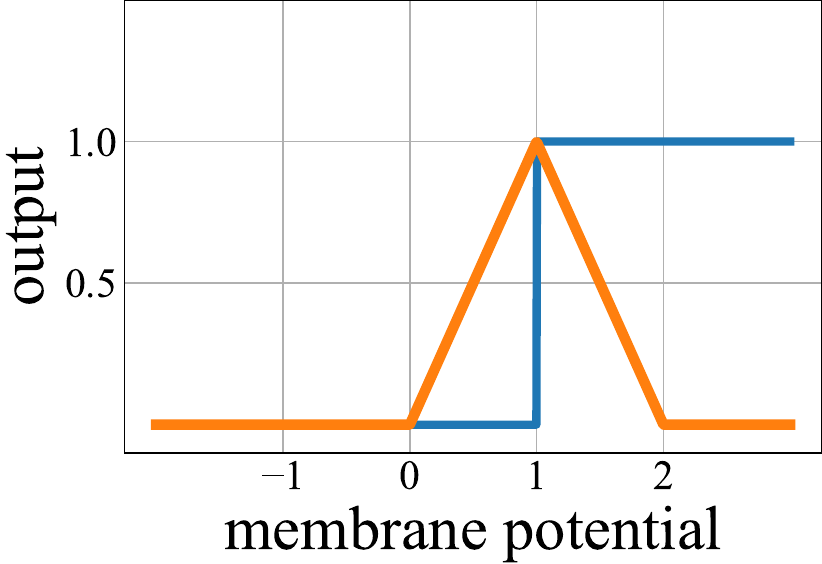}\\
    \subcaption{EENC}
    \label{fig:surrogate-gradients-eenc}
\end{minipage}
\hfill
\begin{minipage}[b]{0.3\linewidth}
    \centering
    \includegraphics[keepaspectratio,width=\linewidth]{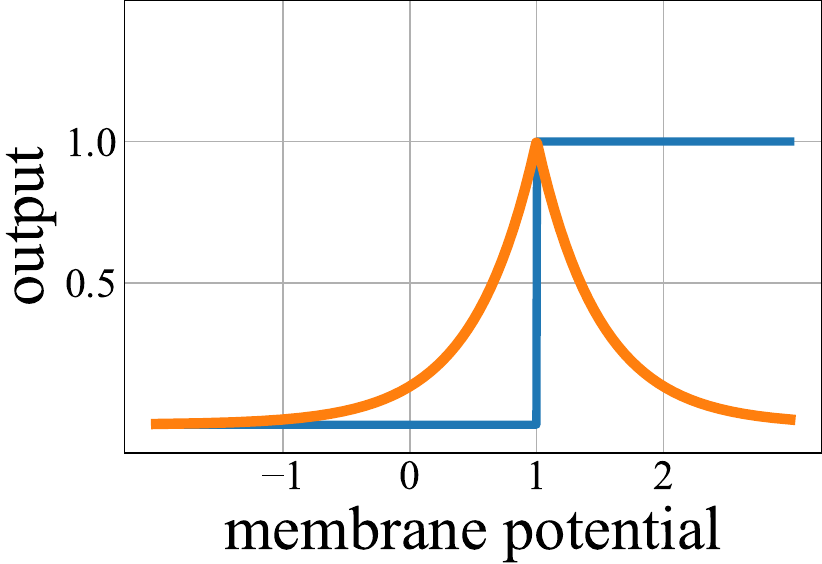}\\
    \subcaption{SLAYER}
    \label{fig:surrogate-gradients-slayer}
\end{minipage}
\hfill
\begin{minipage}[b]{0.3\linewidth}
    \centering
    \includegraphics[keepaspectratio,width=\linewidth]{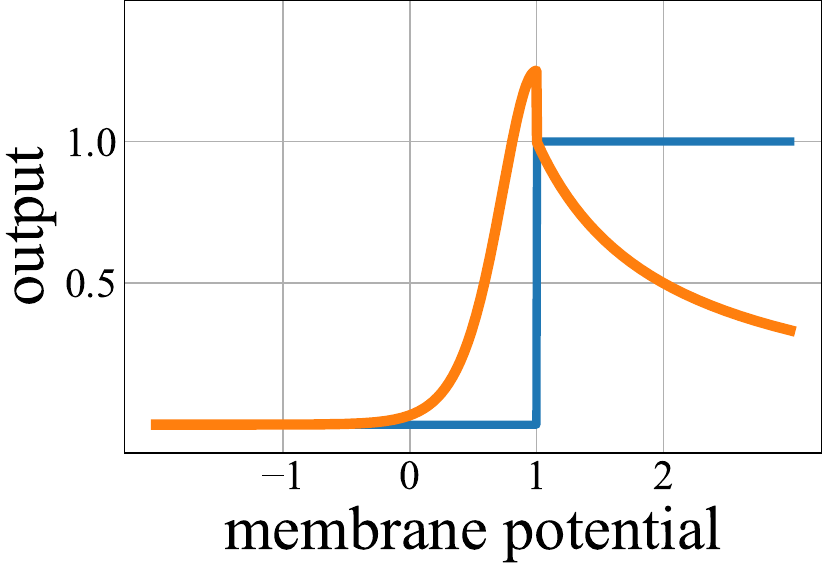}\\
    \subcaption{Ours (Eq.~\ref{eqn:ssn-surrogate-gradient})}
    \label{fig:surrogate-gradients-s3nn}
\end{minipage}
\caption{Surrogate gradients for the Heaviside function used in our experiments: The blue line represents forward propagation while the orange line represents backpropagation (surrogate gradient). }
\label{fig:surrogate-gradients}
\end{figure}

With SNNs as the starting point of our idea, its most crucial aspect is found in the surrogate gradient (Fig.~\ref{fig:surrogate-gradients}).
We experimentally show that our surrogate gradient, which is derived from the multi-time step SNN~\citep{zhang2020temporal}, achieves a superior balance between accuracy and energy efficiency than existing methods.

\paragraph{Main contribution}
We developed a neuron model for SNNs with the time step $\to 1$.
Using the proposed neuron model, we can train the neural networks with high energy efficiency and accuracy close to that of the full-precision network~\footnote{The full-precision network means the ANN counterpart of the referred network.} without the cost of \Erase{multiple}\Add{multi}-time steps.
We call this neuron model a {\it single-step spiking neuron} and use the term {\it single-step spiking neural network} ({\it S$^3$NN}) to describe any neural network using this neuron.

\paragraph{Paper organization} 
In \S~\ref{sec:related-work}, we summarize the related work and clarify the contribution of this study.
In \S~\ref{sec:single-step-neuron}, we introduce neuron models used by SNNs and provide a detailed explanation of how to achieve a limit of time steps $\to 1$.
In \S~\ref{sec:single-step-neural-networks}, we present an example of S$^3$NN using the single-step spiking neuron.
Finally, in \S~\ref{sec:experiment}, we demonstrate the practicality of the proposed model through three different static image classification tasks.

\section{Related Work}
\label{sec:related-work}

\subsection{Three Generations of Neural Networks}
\label{sec:first-generation-neural-networks}
Neural networks can be classified into three generations~\citep{maass1997networks,wang2020supervised}. The first-generation models use McCulloch--Pitts neurons~\citep{mcculloch1943logical} whose outputs are the binary $\{0,1\}$ by thresholding.
The second-generation models use the differentiable activation function and output the real-valued.
The third-generation models are SNNs that use biologically plausible spiking neurons.
Inputs and outputs of the spiking neurons are $\{0, 1\}$.
They can retain information on time-series spike trains.
However, we consider SNNs with a limit of time steps $\to 1$.
In this case, the spiking neuron becomes the step activation function, and the function for time-series information vanishes (\S~\ref{sec:single-step-neuron}).
Therefore, the derived neuron becomes the same as the first-generation neurons.
The difference is the surrogate gradient.
Our contribution can also be viewed as deepening the first generation model by the surrogate gradient derived naturally from the multi-time step SNN~\citep{zhang2020temporal}.
Deep belief networks~\citep{hinton2006fast} and deep Boltzmann machines~\citep{salakhutdinov2009deep} also use the $\{0,1\}$ information in the hidden layers.
However, unlike our deterministic model, they are the generative models by the probabilistic activation function.
Meanwhile, they are out of scope since the concepts are significantly different.

\subsection{Binary Neural Networks}
\label{sec:surrogate-gradient-for-binary-neural-networks}
Typical ANNs that represent weights and inputs/outputs of each layer by real-valued have to compute many MAC operations.
However, BNNs binarize weights and inputs/outputs and replace MAC operations with more efficient xnor/popcount operations to reduce the computational cost~\citep{rastegari2016xnor,wang2020sparsity}.
Such low computational cost techniques can also be applied to SNNs~\citep{roy2019scaling,lu2020exploring,kheradpisheh2020bs4nn}.
However, our proposed method treats the weights as real-valued, as in ordinary SNNs, from the viewpoint of accuracy. 

In BNNs, each layer outputs ${\pm 1}$ by the non-differential step functions (a Heaviside function where the firing threshold is zero), and straight-through estimators (STEs)~\citep{bengio2013estimating,yin2019understanding,shekhovtsov2021reintroducing} are widely used as surrogate gradients~\citep{hubara2016binarized,rastegari2016xnor} (Fig.~\ref{fig:surrogate-gradients-ste-b}).
Generally, STE defines the gradient of a step function as one if an input is in $[-1, 1]$ or zero otherwise. \citet{wang2020sparsity} reported that making them as adaptive and binarizing like SNNs ({\it i.e.}, $\{0, 1\}$) improves the inference accuracy and sparseness~\footnote{Sparseness, as used here, refers to the percentage of inactive neurons, with higher values being more desirable.} (Fig.~\ref{fig:surrogate-gradients-sibnn}).
Various STEs have been proposed~\citep{yuan2021comprehensive}, and it is possible to replace them with the surrogate gradient of single-step spiking neurons.
However, our experiments demonstrate that the surrogate gradient derived from the multi-time step SNN~\citep{zhang2020temporal} is more effective overall inference accuracy and energy efficiency. 

\subsection{Spiking Neural Networks}
\label{sec:surrogate-gradient-for-siking-neural-networks}
Since inactive neurons do not contribute to subsequent operations, we can reduce the computational cost and power consumption by only using activated ones.
SNNs realize them by setting the inputs/outputs of each layer to $\{0, 1\}$.
However, SNNs are more difficult to train than ANNs and BNNs because they have time-series states in addition to a non-differentiable spike function.
Among several training approaches, some methods assign a symmetric surrogate gradient to all \Erase{spike times}\Add{time steps}~\citep{zenke2018superspike,shrestha2018slayer,neftci2019surrogate} (${\it e.g.}$, see Figs.~\ref{fig:surrogate-gradients-eenc} and \ref{fig:surrogate-gradients-slayer}).
These methods reduce the accuracy since it is too simple and does not consider the spike timing.
To solve this problem, \citet{zhang2020temporal} proposed assigning an imbalanced surrogate gradient based on the spike timing.
They improved the inference accuracy and significantly reduced the number of required time steps using this method.
Since this method can achieve fewer time steps, we assume that the \Erase{multi-time step}surrogate gradient \Add{depending on the spike timing} is suitable \Add{also} for limiting time step $\to 1$.
The derived surrogate gradient is shown in Fig.~\ref{fig:surrogate-gradients-s3nn}.

In addition to the surrogate gradients, there is another approach for training SNNs.
This approach reuses the trained ANN's parameters while converting the activation function for the spike function~\citep{diehl2015fast,neil2016learning,han2020rmp,ding2021optimal,pmlr-v139-li21d,kim2022privatesnn}.
However, since the accuracy of this method tends to be proportional to the number of time steps, it is not easy to obtain high accuracy when the number of time steps is small (especially when the time step $=1$).
Meanwhile, \citet{chowdhury2021one} and \citet{severa2018whetstone} proposed realizing the single-time step SNNs by iteratively reducing the number of time steps after the conversion and by gradually changing the activation function to a more sharpened function, respectively.
Using \Erase{this method}\Add{these methods}, the trained model becomes the same as our trained one.
However, our approach does not require the trained ANN and iterative time step reduction.
\Erase{Thus, we can avoid their iteration and the deterioration of accuracy caused by the conversion.}

\section{Single-Step Spiking Neuron}
\label{sec:single-step-neuron}

Based on the spiking neuron model widely used in SNNs, we take the time step $\to 1$ limit and establish a new neuron model, {\it{i.e., a single-step spiking neuron}}.
For a feed-forward neuron model, it is the same as the step activation function widely used in the history of neural networks.
We demonstrate that this neuron model possesses a nontrivial backward mechanism, {\it i.e.,} surrogate gradient.

\subsection{Forward Pass for Base Spiking Neuron Model}
\label{sec:base-forward}

We adopt the leaky integrate-and-fire (LIF) neurons to construct SNNs.
LIF neurons for continuous or discrete time systems can be modeled as follows.
First, let denote $t \in \boldsymbol{R}$ as the continuous time, $l \in \boldsymbol{Z}_{\geq 0}$ as the layer, and $d_l$ as the number of units in $l$-th layer.
Then, the time response of the membrane potential $\boldsymbol{u}^{(l)}(t) \in \boldsymbol{R}^{d_{l}}$ and the postsynaptic current $\boldsymbol{a}^{(l)}(t) \in \boldsymbol{R}^{d_{l}}$ are modeled using the following equations:
\begin{align}
\label{eqn:membrane-c}
\tau_m \frac{d\boldsymbol{u}^{(l)}(t)}{dt}
&= -\boldsymbol{u}^{(l)}(t) + \boldsymbol{W}^{(l)}\boldsymbol{a}^{(l-1)}(t) + \boldsymbol{\eta}^{(l)}(t), \\
{}
\label{eqn:psc-c}
\tau_s \frac{d \boldsymbol{a}^{(l)}(t)}{d t}
&= -\boldsymbol{a}^{(l)}(t) + \boldsymbol{s}^{(l)}(t),
\end{align}
where $\boldsymbol{W}^{(l)} \in \boldsymbol{R}^{d_{l}\times d_{l-1}}$ is the strength of synapse connections (weight matrix) and $\tau_m, \tau_s > 0$ are the time constant of the membrane and postsynaptic, respectively\footnote{We obtain the binary spiking neural network if elements of $\boldsymbol{W}^{(l)}$ are restricted to the binary.
}.
Furthermore, $\boldsymbol{s}^{(l)}(t) \in \{0, 1\}^{d_l}$ denotes a spike train, which is expressed by the Heaviside function $H$ as
\begin{align}
\label{eqn:spike-c}
s_i^{(l)}(t)
= H\left(u_i^{(l)}(t)-u_{\rm th}\right)
= \left\{
\begin{array}{ll}
1 & \left(u_i^{(l)}(t)\ge u_{\rm th}\right) \\
0 & \left(u_i^{(l)}(t)< u_{\rm th}\right),
\end{array}
\right.
\end{align}
where $u_{\rm th} \in \boldsymbol{R}$ is the firing threshold.
Therefore, $s_i^{(l)}(t) = 1$ means that the $i$-th neuron in the $l$-th layer fires at time $t$.
$\boldsymbol{\eta}$ in Eq.~\ref{eqn:membrane-c} denotes the reset mechanism, which changes a membrane potential $u_i^{(l)}$ to the resetting membrane potential $u_{\text{reset}} = 0$ when its corresponding neuron fires.
$\boldsymbol{\eta}$ may also model the refractory period, in which a membrane potential of a fired neuron cannot be changed.
Since we consider the case of the time step $\to$ 1, we ignore the membrane potential responses after firing its corresponding neuron; thus, the reset mechanism and refractory period will not be discussed.

The synapse model equation in Eq.~\ref{eqn:psc-c} can be expressed using the spike response kernel $\varepsilon$ as
\begin{align}
\label{eqn:spike-response-kernel}
\boldsymbol{a}^{(l)}(t)
= \varepsilon * \boldsymbol{s}^{(l)}(t).
\end{align}
Here, we use the single exponential model as $\varepsilon$:
\begin{align}
\label{eqn:single-exponential-model}
\varepsilon(t)
= \frac{1}{\tau_s} \exp\left(-\frac{t}{\tau_s}\right).
\end{align}

A spike train $s_i^{(l)}$ can be detected by firing times $\bar{t}$ that satisfy $s_i^{(l)}(\bar{t})=1$ as
\begin{align}
s_i^{(l)}
\sim \left\{\bar{t}_{i_1}^{(l)}, \ldots, \bar{t}_{i_n}^{(l)}\right\}.
\end{align}
Here, each $\bar{t}$ is locally differentiable with respect to $u_i^{(l)}$~\citep{bohte2002error}:
\begin{align}
\bar{t}^{(l)}
&= \bar{t}^{(l)}\left(u_i^{(l)}\right), \\
{}
\label{eqn:localization}
\frac{\partial \bar{t}^{(l)}}{\partial u_i^{(l)}}\left(u_i^{(l)}\right)
&= -\frac{1}{du_i^{(l)}/dt}.
\end{align}
%

Next, we apply the forward Euler method to the neuron model for continuous time system in Eqs.~\ref{eqn:membrane-c} and \ref{eqn:psc-c}.
Then, we can obtain the neuron model for discrete time system as
\begin{align}
\label{eqn:membrane-d}
\boldsymbol{u}^{(l)}[t]
&= \left(1-\frac{1}{\tau_m}\right)\boldsymbol{u}^{(l)}[t-\Delta t]
+ \boldsymbol{W}^{(l)}\boldsymbol{a}^{(l-1)}[t] + \boldsymbol{\eta}^{(l)}[t], \\
{}
\label{eqn:psc-d}
\boldsymbol{a}^{(l)}[t]
&= \left(1-\frac{1}{\tau_s}\right)\boldsymbol{a}^{(l)}[t-\Delta t] + \boldsymbol{s}^{(l)}[t],
\end{align}
where $\Delta t > 0$ is the finite time period and some variables have been redefined, such as $\tau_* / \Delta t \to \tau_*$, $\boldsymbol{W} / \Delta t \to \boldsymbol{W}$.
Similarly, a spike train is defined as continuous system in Eq.~\ref{eqn:spike-c}:
\begin{align}
\label{eqn:spike-d}
s_i^{(l)}[t]
&= H\left(u_i^{(l)}[t]-u_{\rm th}\right)
= \left\{
\begin{array}{ll}
1 & \left(u_i^{(l)}[t]\ge u_{\rm th}\right) \\
0 & \left(u_i^{(l)}[t] < u_{\rm th}\right).
\end{array}
\right.
\end{align}

\subsection{Backward Pass for Base Spiking Neuron Model}
\label{sec:base-backward}

We can process various parts of the backpropagation of SNNs in the same way as the backpropagation through time used in recurrent neural networks~\citep{werbos1990backpropagation}.
However, the postsynaptic current $\boldsymbol{a}^{(l)}$ in SNNs is not differentiable since it follows the spike firing $\boldsymbol{s}^{(l)}$.
Thus, we need to establish some model for $\partial \boldsymbol{a}^{(l)}[t_k]/\partial \boldsymbol{u}^{(l)}[t_m]$ $(t_k = k \Delta t, \, k \in \boldsymbol{Z}_{\geq 0}, \, k\geq m)$, {\it i.e.}, some surrogate gradient.

To be practical about this argument, we analyze the surrogate gradient called {\it temporal spike sequence learning via backpropagation} ({\it TSSL-BP}) which can train SNNs in a few time steps~\citep{zhang2020temporal}:
\begin{align}
\label{eqn:tsslbp-full}
\frac{\partial a_{i}^{(l)}\left[t_{k}\right]}{\partial u_{i}^{(l)}\left[t_{m}\right]}
= \phi_{i}^{(l)}\left(t_{k}, t_{m}\right)
=\phi_{i}^{(l)\langle 1\rangle}\left(t_{k}, t_{m}\right)
+\phi_{i}^{(l)\langle 2\rangle}\left(t_{k}, t_{m}\right).
\end{align}
The TSSL-BP distinguishes two patterns for the spike propagation: $\phi^{\langle 1\rangle}, \phi^{\langle 2\rangle}$.
Both patterns describe the effect of spike firing at $t_m$ on the postsynaptic current $\boldsymbol{a}^{(l)}[t_k]$ $(k \geq m)$.
However, $\phi^{\langle 1\rangle}$ only models a direct influence whereas $\phi^{\langle 2\rangle}$ models an indirect influence through the reset mechanism $\eta$ when the neuron also fire at $t_p$ $(m<p<k)$.
In this study, we consider the pattern $\phi^{\langle 1\rangle}$ where a single spike firing causes a non-zero value:
\begin{align}
\label{eqn:tsslbp-first}
\phi_{i}^{(l)\langle 1\rangle}\left(t_{k}, t_{m}\right)
=\frac{\partial a_{i}^{(l)}\left[t_{k}\right]}{\partial t_{m}} \frac{\partial t_{m}}{\partial u_{i}^{(l)}\left[t_{m}\right]}.
\end{align}
The other formula $\phi^{\langle 2\rangle}$ is always zero until a subsequent spike firing occurs.
For further details on $\phi^{\langle 2\rangle}$, refer to the original paper~\citep{zhang2020temporal}.

\subsection{Derivation of Spiking Neuron for Limiting Time Step $\to$ 1}
\label{sec:derivation-ssn}

A spiking neuron for time step $\to$ 1 can be deduced by eliminating temporal dependence on the LIF neuron model in the neighborhood of a certain time $t$; we set $t=0$ for sake of simplicity.
When we focus on the transition of a single-time step at $t=-\Delta t \to 0$ in the LIF neuron model, we first assume that the discrete time neuron model (Eqs.~\ref{eqn:membrane-d}--\ref{eqn:spike-d}) is in its initial state, $\boldsymbol{u}^{(l)}[-\Delta t] = \mathbf{0}, \boldsymbol{a}^{(l)}[-\Delta t] = \mathbf{0}$.
Our focus is only on what occurs while a unit of time $\Delta t$ passes, ignoring everything afterward.
This allows us to exclude the reset mechanism $\boldsymbol{\eta}$ and obtain $\boldsymbol{u}^{(l)}[0] = \boldsymbol{W}^{(l)}\boldsymbol{a}^{(l-1)}[0]$, $\boldsymbol{a}^{(l)}[0] = \boldsymbol{s}^{(l)}[0]$.
Thus, we define the resultant model as the feed-forward single-step spiking neuron model:
\begin{align}
\label{eqn:membrane-ssn}
\boldsymbol{u}^{(l)}
&= \boldsymbol{W}^{(l)}\boldsymbol{a}^{(l-1)}, \\
{}
\label{eqn:psc-ssn}
\boldsymbol{a}^{(l)}
&= \boldsymbol{s}^{(l)}, \\
{}
\label{eqn:spike-ssn}
s_i^{(l)}
= H\left(u_i^{(l)}-u_{\rm th}\right)
&= \left\{
\begin{array}{ll}
1 & \left(u_i^{(l)}\ge u_{\rm th}\right) \\
0 & \left(u_i^{(l)}< u_{\rm th}\right).
\end{array}
\right.
\end{align}

Next, we deduce the surrogate gradient of the single-step spiking neuron.
Here we assume TSSL-BP (Eq.~\ref{eqn:tsslbp-full}) as the surrogate gradient for the LIF neuron.
Therefore, the surrogate gradient of the single-step spiking neuron corresponds to Eq.~\ref{eqn:tsslbp-full} for the condition $t_k=t_m=0$.
If $\bar{t}_i^{(l)}=\emptyset$, Eq.~\ref{eqn:tsslbp-full} vanishes trivially.
Then, we set $\bar{t}_i^{(l)}=\{0\}$.
By substituting
Eq.~\ref{eqn:tsslbp-first} and $\phi_{i}^{(l)\langle 2\rangle}\left(0, 0\right) = 0$ into Eq.~\ref{eqn:tsslbp-full}, we obtain the following equation:
\begin{align}
\label{eqn:derivation-ssn}
\frac{\partial a_{i}^{(l)}\left[0\right]}{\partial u_{i}^{(l)}\left[0\right]}
&=\phi_{i}^{(l)\langle 1\rangle}\left(0, 0\right)
  +\phi_{i}^{(l)\langle 2\rangle}\left(0, 0\right)
{} \nonumber \\
&
=\left.\frac{\partial a_{i}^{(l)}\left[t\right]}{\partial \bar{t}}\right|_{t=0, \bar{t}=0}
\left.\frac{\partial \bar{t}}{\partial u_{i}^{(l)}\left[t\right]}\right|_{t=0, \bar{t}=0}.
\end{align}
%
Here, in the same way as ~\citet[Eq.~12]{zhang2020temporal}, Eq.~\ref{eqn:spike-response-kernel} implies followings:
\begin{align}
\left.\frac{\partial a_{i}^{(l)}\left[t\right]}{\partial \bar{t}}\right|_{t=0, \bar{t}=0}
= \left.\frac{d \left(\varepsilon * s_i^{(l)}[\bar{t}=0]\right)}{d t}\right|_{t=0}
= -\frac{1}{\tau_s}s_{i}^{(l)}[0]
= -\frac{1}{\tau_s},
\end{align}
where the second equality follows from Eq.~\ref{eqn:single-exponential-model} with the renormalization multiplier $\tau_s$ for the discrete time system (Eq.~\ref{eqn:psc-d}),
and the last equality follows from $\bar{t}_i^{(l)}=\{0\}$.
Moreover, Eq.~\ref{eqn:localization} implies followings:
\begin{align}
\left.\frac{\partial \bar{t}}{\partial u_{i}^{(l)}\left[0\right]}\right|_{\bar{t}=0}
= -\frac{1}{\left.du_{i}^{(l)}/dt\right|_{t=0}}
= -\frac{1}{\sum_j W_{ij}^{(l)}a_{j}^{(l-1)}[0]},
\end{align}
where the last equality follows from Eq.~\ref{eqn:membrane-d} and assumptions that $\boldsymbol{u}^{(l)}[-\Delta t] = \mathbf{0}$ and $\boldsymbol{\eta}=0$.
By substituting Eq.~\ref{eqn:membrane-ssn} into Eq.~\ref{eqn:derivation-ssn}, we can define the surrogate gradient of the single-step spiking neuron at spike firing by the following equation:
\begin{align}
\frac{\partial a_{i}^{(l)}}{\partial u_{i}^{(l)}}\left(u_{i}^{(l)}\right)
= \frac{1}{\tau_s} \frac{1}{u_{i}^{(l)}} \quad
\left(u_{i}^{(l)} \geq u_{\rm th}\right).
\end{align}

However, TSSL-BP (Eq.~\ref{eqn:tsslbp-full}) cannot return a gradient unless spike firings occur and cannot train corresponding neurons.
Thus, we employ the warm-up mechanism~\citep{zhang2020temporal} that allows us to provide a nontrivial surrogate gradient even when a neuron is not firing and prevent the problem of the dead neuron.
In this paper, as in~\citep{zhang2020temporal}, the warm-up mechanism gives a neuron which is not firing the derivative of the scaled sigmoid function given as follows:
\begin{align}
\sigma_\alpha\left(u_{i}^{(l)}\right)
&= \frac{1}{1 + \exp\left(\left(-u_{i}^{(l)}+u_{\rm th}\right)/\alpha\right)},
{} \\
\frac{\partial \sigma_\alpha}{\partial u_{i}^{(l)}}\left(u_{i}^{(l)}\right)
&= \frac{1}{\alpha} \sigma_\alpha\left(u_{i}^{(l)}\right) \left(1 - \sigma_\alpha\left(u_{i}^{(l)}\right)\right),
\end{align}
where $\sigma_\alpha$ is the scaled sigmoid function and $\alpha > 0$ is its hyperparameter.
Note that the choice for the warm-up function is unsettled, and there are other candidates, such as the derivative of the fast sigmoid in \citet{zenke2018superspike} and the exponentially decaying function in \citet{shrestha2018slayer}.

In summary, the surrogate gradient of a single-step spiking neuron is defined as follows:
\begin{align}
\label{eqn:ssn-surrogate-gradient}
\frac{\partial a_{i}^{(l)}}{\partial u_{i}^{(l)}}\left(u_{i}^{(l)}\right)
= \left\{
\begin{array}{ll}
\frac{1}{\tau_s} \frac{1}{u_{i}^{(l)}} & \left(u_{i}^{(l)} \geq u_{\rm th}\right) \\
\frac{1}{\alpha} \sigma_\alpha\left(u_{i}^{(l)}\right) \left(1 - \sigma_\alpha\left(u_{i}^{(l)}\right)\right) & \left(u_{i}^{(l)} < u_{\rm th}\right)
\end{array}
\right..
\end{align}
This surrogate gradient is presented in Fig.~\ref{fig:surrogate-gradients-s3nn}, where the parameters in Eq.~\ref{eqn:ssn-surrogate-gradient} are set to $u_{\rm th} = 1.0$, $\tau_s = 1.0$, and $\alpha = 0.2$.
Note that the single-step spiking neuron has no time-dependency and is not the LIF neuron anymore.

%

\section{Single-Step Spiking Neural Networks}
\label{sec:single-step-neural-networks}
\begin{figure}[t]
\begin{minipage}[b]{0.32\linewidth}
    \centering
    \includegraphics[keepaspectratio,height=60mm]{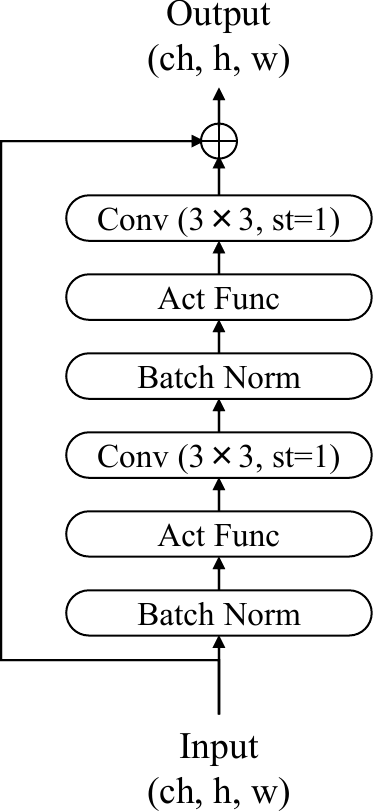}
    \subcaption{Residual module-A}
    \label{fig:architecture-res-a}
\end{minipage}
\hfill
\begin{minipage}[b]{0.32\linewidth}
    \centering
    \includegraphics[keepaspectratio,height=60mm]{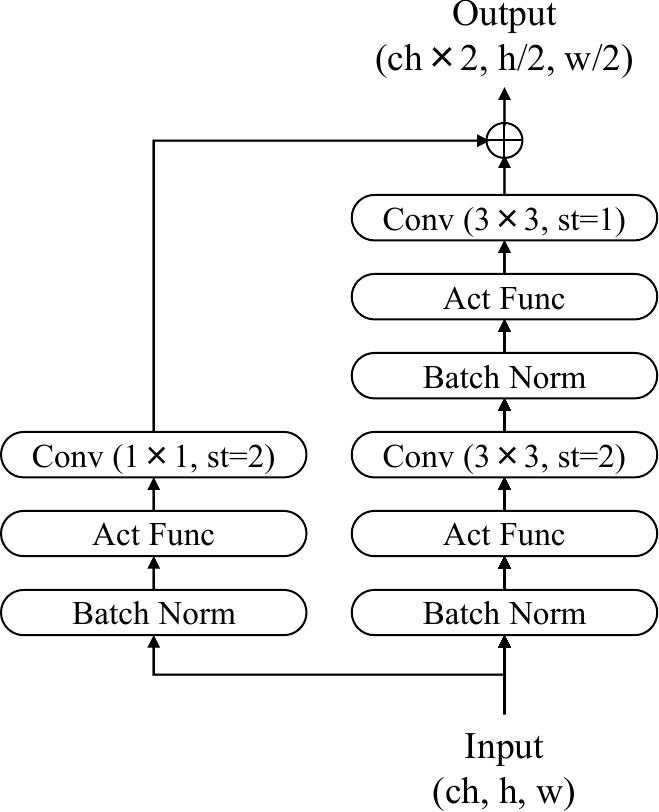}
    \subcaption{Residual module-B}
    \label{fig:architecture-res-b}
\end{minipage}
\hfill
\begin{minipage}[b]{0.32\linewidth}
    \centering
    \includegraphics[keepaspectratio,height=60mm]{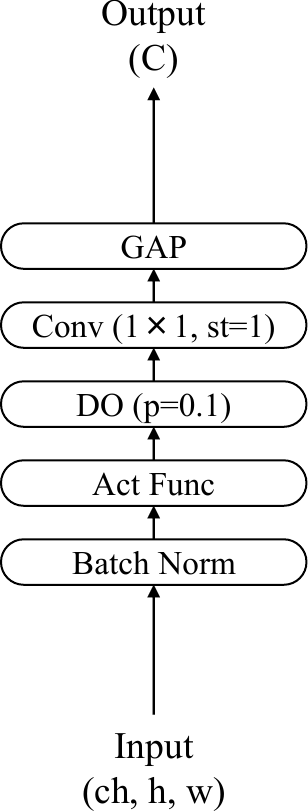}
    \subcaption{Output layer}
    \label{fig:architecture-out}
\end{minipage}
\caption{Diagram of each module that constitutes the network:
Conv refers to the convolutional layer, Act Func refers to the activation function, Batch Norm refers to BN Layer, GAP refers to the global average pooling, and DO refers to dropout.
The convolutional layer is given the kernel and stride size, while the dropout is given a dropout ratio.
As input/output, ch refers to channel size; h and w refer to the height and width of the feature map, respectively; C refers to the number of classes.}
\label{fig:architecture}
\end{figure}

Using the single-step spiking neuron obtained in the previous section, we propose an S$^3$NN that accurately classifies static images with a low computational cost.
The proposed network is built upon a pre-activation ResNet (PreActResNet)~\citep{he2016identity}, {\it i.e.}, a stack of pre-activation type residual modules (Fig.~\ref{fig:architecture}).
In this section, we explain how to handle the encoding of the static images, loss function, and batch normalization (BN) layers.
Note that the postsynaptic current with the surrogate gradient for the single-step spiking neuron in Eqs.~\ref{eqn:psc-ssn} and \ref{eqn:ssn-surrogate-gradient} is called activation function below.

\subsection{Encoding}
\label{sec:input-layer}
Some SNNs encode input data as spike trains~\citep{shrestha2018slayer}, whereas some BNNs encode them as binary~\citep{hubara2016binarized}.
In this paper, we feed the static image $\boldsymbol{a}^{(l=0)} \in \boldsymbol{R}^{d_0}$ to our input layer and convert it to the binary map $\boldsymbol{s}^{(1)} =  H(\boldsymbol{u}^{(1)}-u_{\rm th})$, where $\boldsymbol{u}^{(1)} = \boldsymbol{W}^{(1)}\boldsymbol{a}^{(0)}$.
In the context of SNNs, this direct image input is called direct encoding~\citep{rueckauer2017conversion,lu2020exploring,rathi2020diet,shi2020accurate,zheng2021going}.
The difference to the SNN's direct encoding is that our encoding outputs the binary map, not the spike train.
Note that the following layers take binary maps $\boldsymbol{s}^{(l\neq 0)}$ as their input data.

\subsection{Loss function}
\label{sec:output-layer}
Several methods in SNNs usually convert the output data into a spike format through the spiking neuron to calculate the loss function by counting the number of spikes~\citep{shrestha2018slayer} or measuring their van Rossum distance between the desired and actual spikes~\citep{van2001novel}.
However, converting the output data into spikes may result in information loss.
Instead, we directly set the output of the S$^3$NN as the logits $\boldsymbol{y} = (y_c)_{c=1}^{C} \in \boldsymbol{R}^C$ (Fig.~\ref{fig:architecture-out}). 
As the loss function, we use the softmax-cross entropy function with the label smoothing~\citep{szegedy2016rethinking} to prevent overfitting as follows:
\begin{align}
\label{eqn:loss-function}
{\rm {Loss}}(\boldsymbol{y}, \boldsymbol{t})
= - \sum_{c} \left(t_c(1-\epsilon) + \frac{\epsilon}{C}\right) \log \frac{\exp(y_{c})}{\sum_{k} \exp(y_{k})},
\end{align}
where $C$ represents the number of classes, $\boldsymbol{t} = (t_c)_{c=1}^{C} \in \{0, 1\}^C$ represents the one-hot encoded label, and $\epsilon > 0$ represents the smoothing parameter.

\subsection{Batch Normalization Layer}
\label{sec:batch-normalization-layer}
As adopted in many studies on ANNs, in this paper, the BN layer is also positioned immediately before activation in training phase.
However, during inference phase, computational costs for the BN layer can be reduced by eliminating the BN layer and re-normalizing the other parameters appropriately.
In this work, as in \citep{qiao2020stbnn}, we can remove the BN layer of the S$^3$NN by merging parameters of the BN layer into the firing threshold of neurons.
In fact, by expressing the affine transformation with respect to $u$ as $\bar{u} = \gamma \cdot u + \beta$ $(\gamma > 0, \beta \in R)$, it follows that
\begin{align}
\bar{u} = \gamma \cdot u + \beta \gtrless u_{\rm th}
\Longleftrightarrow u \gtrless u'_{\rm th}=\frac{u_{\rm th} - \beta}{\gamma}.
\end{align}
In other words, the BN layer represented as an affine transformation can be completely absorbed into neurons by making the firing threshold for each neuron variable.
Note that this idea cannot be applied to SNNs with spiking neurons that have temporal states.

\section{Experiment}
\label{sec:experiment}
Based on the image classification task (Fashion-MNIST~\citep{xiao2017fashion}, CIFAR-10, 100~\citep{krizhevsky2009learning}, TinyImageNet~\citep{Le2015TinyIV}), we demonstrate that the proposed S$^3$NNs has a low degradation of inference accuracy from the full-precision networks.
In particular, the single-step spiking neurons using TSSL-BP~\citep{zhang2020temporal} are highly accurate and show the energy efficiency compared to other surrogate gradients.
Additionally, they exhibit a superior balance between inference accuracy and latency than existing BNNs and SNNs. 
We conducted experiments on an NVIDIA GeForce RTX 3090.

\subsection{Experimental Setup}
\label{sec:experimental-setups}
The network architecture of the proposed S$^3$NN is a PreActResNet as described in \S \ref{sec:single-step-neural-networks}, where the pre-activation type residual modules (Fig.~\ref{fig:architecture}) are stacked (these are called ResNet18 or ResNet106 based on the number of layers).
The full-precision network is the baseline for comparing inference accuracy and energy efficiency.
It replaces the above S$^3$NN activation function with the ReLU function.
Adam or SGD, whichever yields better inference accuracy, is used as the optimizer in each training instance.
Regularization methods such as dropout, label smoothing, random augmentation, and test time augmentation are employed.
Refer to \ref{app:hyper-parameters} for information on various hyperparameters and technical details.

\subsection{Classification Accuracy}
\label{sec:classification-accuracy}

We first investigated the effectiveness of our surrogate gradient in training the S$^3$NN. Table~\ref{tbl:comparison-cifar10-a} summarizes the results of various surrogate gradients for the image classification task for the CIFAR-10 dataset, where all models are S$^3$NNs though their activation functions are different \Add{(other settings are the same)}.
The ``Method" in Table~\ref{tbl:comparison-cifar10-a} indicates the activation function employed.
The ``ReLU" method corresponds to the full-precision network.
Meanwhile, the other method corresponds to some network, whose activation function is the Heaviside function with the threshold value $u_{\rm th}$ expressed in Eq.~\ref{eqn:spike-ssn} for forward propagation and is each surrogate gradient method for backpropagation.
For the methods used for comparison, the ``STE-b"/``STE-s" method denotes the STE~\citep{bengio2013estimating} used as the BNN baseline;
the ``SiBNN" method denotes the adaptive STE in~\citep{wang2020sparsity};
the ``SLAYER''~\citep{shrestha2018slayer} and ``EENC''~\citep{esser2016cover} methods denote the functions used as the SNN baselines independent of the process of spike firing, unlike the original TSSL-BP.
The difference between ``STE-b" and ``STE-s" is that the range of the Heaviside function of ``STE-b" is not $\{0, 1\}$, but $\{\pm 1\}$. \Add{Namely, ``STE-s'' is a modification of ``STE-b" for BNNs to align the threshold to the SNN setting.}
These surrogate gradients are shown in Fig.\ref{fig:surrogate-gradients} and Table~\ref{tbl:eqs-forward-backward}.
Here, ``Accuracy" expresses the mean value ($\pm$ standard deviation) of the test accuracy in $5$ training runs.

\begin{table}[t]
\caption{Comparisons with some neuron models on the CIFAR-10 dataset}
\label{tbl:comparison-cifar10-a}
\centering
\begin{tabular}{llrrr}
\hline
Architecture & Method & Accuracy [\%] & $E_{\rm {Full}}/E_{\rm {Method}}$ \\ \hline
ResNet18 & STE-b & 89.04$\pm$0.28 & 12.23 \\
ResNet18 & SLAYER & \Add{91.13$\pm$0.14} & \Add{35.51} \\
ResNet18 & EENC & 91.22$\pm$1.21 & 46.92 \\
ResNet18 & STE-s & 91.60$\pm$0.09 & 33.73 \\
ResNet18 & ReLU & 91.67$\pm$0.07 & 28.90 \\
ResNet18 & {\bf {Ours (Eq.~\ref{eqn:ssn-surrogate-gradient})}} & {\bf {93.25$\pm$0.14}} & {\bf {38.96}} \\
ResNet18 & SiBNN & 93.50$\pm$0.20 & 34.75 \\
ResNet106 & EENC & 90.83$\pm$0.18 & 52.42 \\
ResNet106 & SiBNN & 93.98$\pm$0.18 & 38.46 \\
ResNet106 & {\bf {Ours (Eq.~\ref{eqn:ssn-surrogate-gradient})}} & {\bf {94.25$\pm$0.14}} & {\bf {149.33}} \\
ResNet106 & ReLU & 95.51$\pm$0.12 & 1.00 \\
\hline
\end{tabular}
\end{table}

\begin{table}[t]
\caption{Binarization effect on the accuracy for the CIFAR-10 dataset (The bottom line corresponds to the S$^3$NN condition)}
\label{tbl:comparison-cifar10-a-binary}
\centering
\begin{tabular}{lccr}
\hline
Method & Weights & Output & Accuracy [\%]  \\ \hline
Ours (Eq.~\ref{eqn:ssn-surrogate-gradient}) & $\{\pm 1\}$ & $\{\pm 1\}$ & 31.47$\pm$0.82 \\
STE-b & $\{\pm 1\}$ & $\{\pm 1\}$ & 88.97$\pm$0.20 \\
Ours (Eq.~\ref{eqn:ssn-surrogate-gradient}) & $\{\pm 1\}$ & $\{0, 1\}$ & 91.25$\pm$0.14 \\
Ours (Eq.~\ref{eqn:ssn-surrogate-gradient}) & $\boldsymbol{R}$ & $\{0, 1\}$ & 93.25$\pm$0.14 \\
\hline
\end{tabular}
\end{table}

\begin{figure}[t]
\begin{minipage}[b]{0.5\linewidth}
    \centering
    \includegraphics[width=\linewidth]{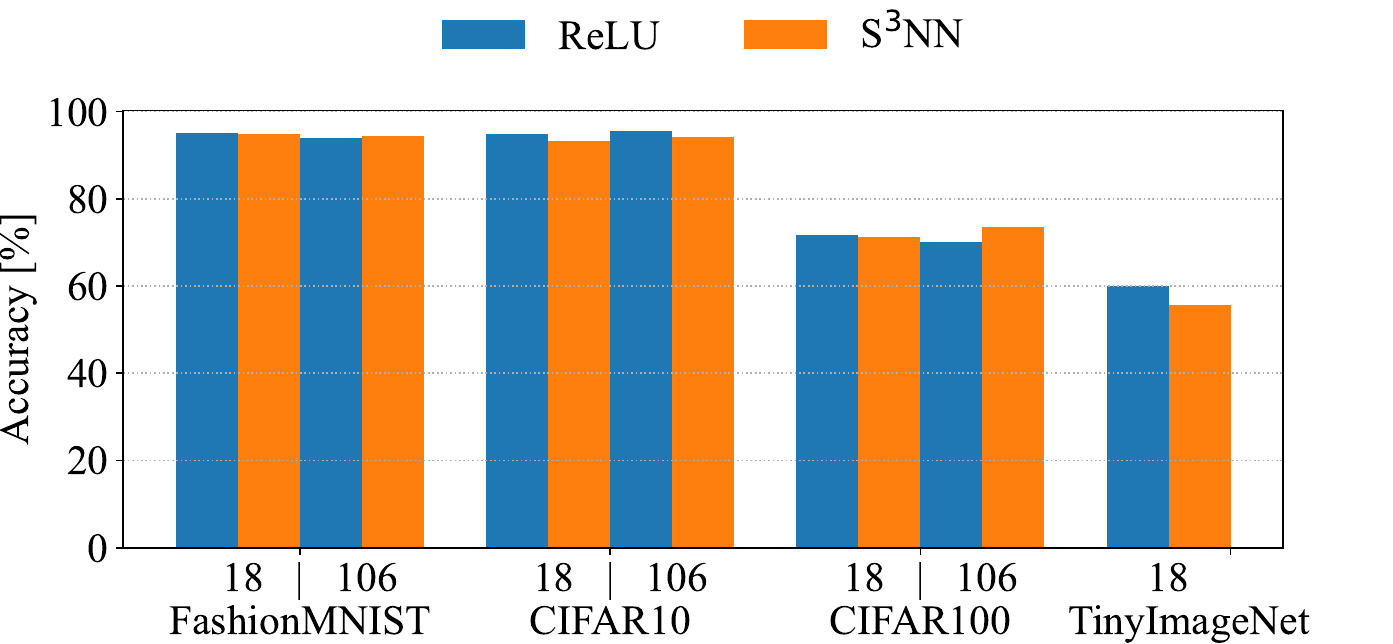}
    \subcaption{Accuracy}
    \label{fig:acc}
\end{minipage}
\hfill
\begin{minipage}[b]{0.5\linewidth}
    \centering
    \includegraphics[width=\linewidth]{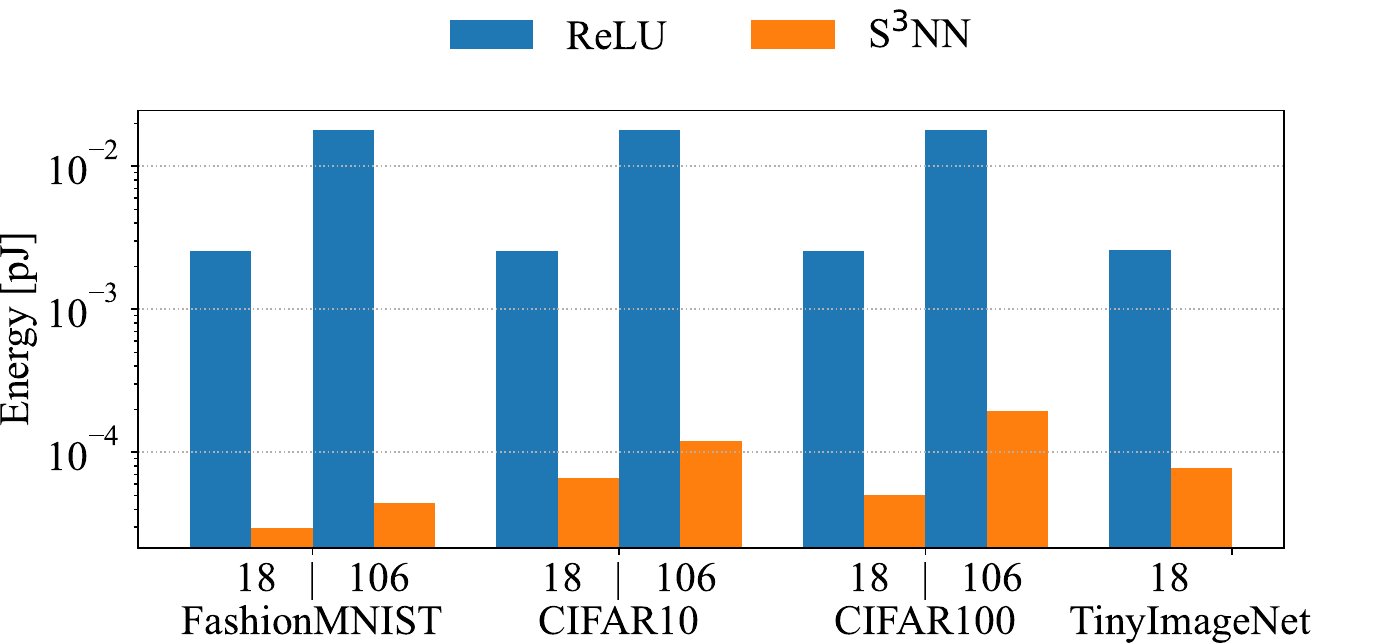}
        \subcaption{Energy consumption}
    \label{fig:energy}
\end{minipage}
\caption{(a) Test accuracy on each dataset and network scale.
(b) Energy consumption in logarithmic scale on each dataset and network scale. The numbers $18$ and $106$ on the $x$-axis denote ResNet18 and ResNet106, respectively.}
\end{figure}

The results on ResNet18 indicate that the proposed method and SiBNN are comparable to full-precision networks in terms of test accuracy.
Additionally, the results on ResNet106 indicate that the proposed method and SiBNN can be appropriately trained, even in deep networks.
In particular, the proposed method shows better scaling than SiBNN.
Note that layer scalability has become nontrivial in SNNs~\citep{zheng2021going}.
As shown in Fig.~\ref{fig:acc}, we can observe similar results on the layer scalability and the comparison between the proposed method and full-precision network also for the Fashion-MNIST and CIFAR-100 datasets\footnote{ResNet106's result for the TinyImageNet was omitted because it could not run on our environment.}.
In addition, surprisingly, our S$^3$NN of ResNet106 outperforms the full-precision network for the CIFAR-100 dataset. This phenomenon also occurs in the other methods~\citep{courbariaux2015binaryconnect,zhang2020temporal,li2021bsnn} and may be due to the redundant expressivity of the full-precision networks. The detail analysis is included in our future work.

The above results are based on the S$^3$NN architecture. To validate the effectiveness of our method in other architectures, we trained several situations shown in Table~\ref{tbl:comparison-cifar10-a-binary}. Note that the condition of $\{ \pm 1 \}$ weights and $\{ \pm 1 \}$ neuron outputs represents BNNs.
We binarized weights according to \citet{rastegari2016xnor} and obtained $\{ \pm 1 \}$ neuron output by a certain linear transformation of each method.
This table shows that the S$^3$NN trained by our surrogate gradient outperformed other conditions. On the other hand, the BNN trained by our surrogate gradient did not work well since ours is deduced from the SNNs function that outputs $\{ 0, 1\}$. Therefore, we should use the S$^3$NN model trained by Eq.~\ref{eqn:ssn-surrogate-gradient} to maintain high accuracy.

\subsection{Energy Efficiency}
\label{sec:energy-efficiency}

The energy efficiency for each method for a full-precision network is estimated according to \citet{lee2020enabling,kim2020revisiting}.
The results are presented in the ``$E_{\rm {Full}}/E_{\rm {Method}}$" column of Table~\ref{tbl:comparison-cifar10-a}.
This metric shows how much each method reduces the energy consumption compared to a full-precision network with the same architecture when implementing asynchronous processing that performs operations only when a spike fires.
The higher the value of $E_{\rm {Full}}/E_{\rm {Method}}$, the more energy-efficient the method.
By definition of the metric, it is highly energy-efficient with a low per-neuron per-data spike firing rate.
For details on the calculation formula, refer to~\ref{app:energy-calculation}.
The proposed method and SiBNN show high test accuracy for the CIFAR-10 dataset (Table~\ref{tbl:comparison-cifar10-a}).
The results on ResNet18 of the proposed method and SiBNN demonstrate a similar level of energy efficiency.
However, from the results on ResNet106, the energy efficiency of the proposed method is superior to that of SiBNN.
Note that, based on Eq.~\ref{eqn:energy-full} (\ref{app:energy-calculation}), the energy consumption of the full-precision ResNet106 is about $6.98$ times larger than that of ResNet18.
Thus, for the $E_{\rm {Full}}/E_{\rm {Method}}$, a larger value is desired for ResNet106 than ResNet18.
Here we can see that the proposed method meets this requirement at a good rate.
Furthermore, as shown in Fig.~\ref{fig:energy}, the energy efficiency also scales at an acceptable rate when the model scale is increased for the Fashion-MNIST and CIFAR-100 datasets.
Although the proposed method for ResNet106 is much larger than ResNet18 in terms of model size, it has advantages over full-precision ResNet18 in terms of the energy efficiency (Fashion-MNIST: 57.97$\times$, CIFAR-10: 21.39$\times$, CIFAR-100: 13.11$\times$).

\begin{table}[h]
\caption{Classification benchmarks for the CIFAR-10 dataset. Superscripts on references are referred to in Fig.~\ref{fig:flops-energy}.
The values for ``Accuracy" are taken from the original papers; the values in parentheses show the accuracy of the full-precision network also described in the original papers.
}
\label{tbl:comparison-cifar10-s}
\centering
\begin{tabular}{lllrl} \hline
{} & Paradigm & Architecture & Time Steps & Accuracy [\%] \\ \hline
\cite{rastegari2016xnor} & BNN & 6Convs.2Linears & 1 & 89.83 \\
\cite{hubara2016binarized} & BNN & 6Convs.2Linears & 1 & 89.85 (89.06) \\
\cite{wang2020sparsity} & BNN & 6Convs.2Linears & 1 & 90.2 \\
\cite{courbariaux2015binaryconnect} & BNN & 6Convs.2Linears & 1 & 91.73 (89.36) \\
\cite{chen2021bnn} & BNN & ReActNet18 & 1 & 92.08 (92.31) \\
\cite{xu2019accurate} & BNN & VGG16-small & 1 & 92.3 (93.6) \\
\cite{shen2020balanced} & BNN & ResNet20 & 1 & 92.46 \\
\cite{wang2019learning} & BNN & VGG-small & 1 & 92.47 (93.2) \\
\cite{esser2016cover} & SNN & 3Convs.8Linears & 1 & 83.41 \\
\cite{severa2018whetstone} & SNN & VGG-like & 1 & 84.67 \\
\cite{kim2022privatesnn} & SNN & VGG16 & 150 & 89.2 (91.6) \\
\cite{zhang2020temporal}$^{(1)}$ & SNN & 5Convs.2Linears & 5 & 91.41 (90.49) \\
\cite{pmlr-v139-yang21n} & SNN & AlexNet & 5 & 91.76 \\
\cite{rathi2019enabling} & SNN & ResNet20 & 250 & 92.22 (93.15) \\
\cite{kundu2021towards} & SNN & VGG16 & 10 & 92.53 (92.97) \\
{\bf {This work}}$^{(2)}$ & S$^3$NN & VGG16 & 1 & {\bf {92.72 (94.39)}} \\
\cite{chowdhury2021one}$^{(3)}$ & SNN & VGG16 & 1 & 93.05 (94.1) \\
\cite{na2022autosnn} & SNN & (NAS) & 8 & 93.15 \\
{\bf {This work}}$^{(4)}$ & S$^3$NN & ResNet18 & 1 & {\bf {93.25 (94.89)}} \\
\cite{deng2020optimal}$^{(5)}$ & SNN & ResNet20 & 128 & 93.56 \\
\cite{han2020rmp} & SNN & VGG16 & 2048 & 93.63 (93.63) \\
\cite{kim2022neural} & SNN & (NAS) & 8 & 94.12 \\
\cite{yan2021near}$^{(6)}$ & SNN & VGG* & 600 & 94.16 (94.2) \\
{\bf {This work}}$^{(7)}$ & S$^3$NN & ResNet106 & 1 & {\bf {94.25 (95.51)}} \\
\cite{pmlr-v139-li21d}$^{(8)}$ & SNN & VGG16 & 64 & 95.14 (95.72) \\
\cite{li2021bsnn} & SNN & ResNet20 & 206 & 95.16 (95.02) \\
\cite{pmlr-v139-li21d}$^{(9)}$ & SNN & ResNet20 & 64 & 95.3 (95.46) \\
\hline
\end{tabular}
\rightline{\Add{{\small (NAS): a neural architecture searched model}}}
\end{table}

\begin{figure}[t]
\begin{minipage}[b]{0.475\linewidth}
    \centering
    \includegraphics[width=\linewidth]{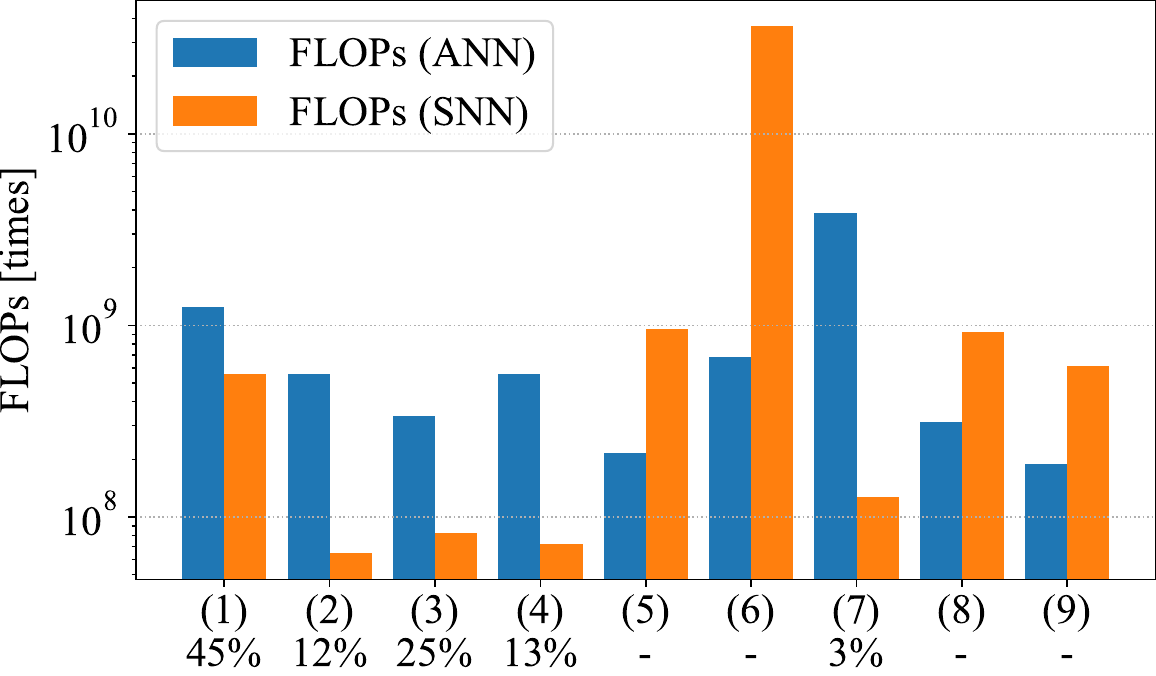}
    \subcaption{FLOPs}
    \label{fig:flops-annsnn}
\end{minipage}
\hfill
\begin{minipage}[b]{0.475\linewidth}
    \centering
    \includegraphics[width=\linewidth]{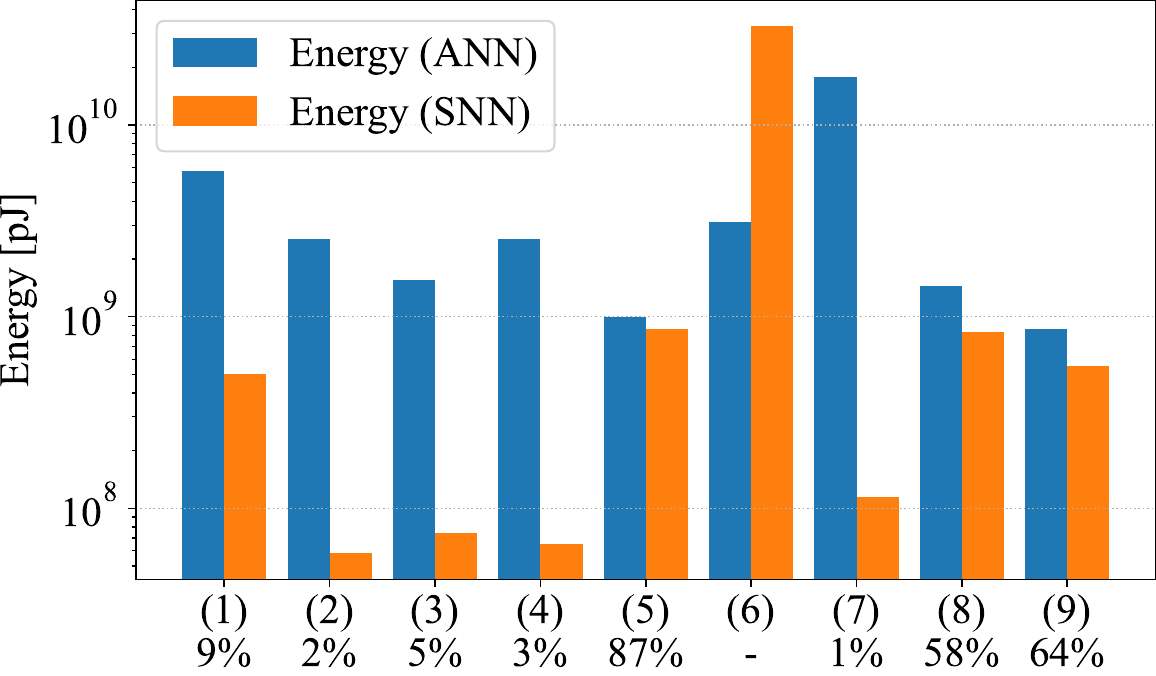}
    \subcaption{Energy consumption}
    \label{fig:energy-annsnn}
\end{minipage}
\caption{FLOPs and energy consumption of each model \Add{(logarithmic notation). The numbers with parentheses on $x$-axis correspond to those listed in Table~\ref{tbl:comparison-cifar10-s}; {\it{e.g.}}, (2), (4) and (7) correspond to our method. The bottom numbers on $x$-axis indicate the reduction percentage from ANNs to SNNs. Note that the values over 100\% are represented by ``-''.}}
\label{fig:flops-energy}
\end{figure}


\subsection{Trade-off between Accuracy and Latency}
\label{sec:tradeoff-between-accuracy-and-latency}
Table~\ref{tbl:comparison-cifar10-s} compares the proposed method and related BNN and SNN methods\Erase{, where the NAS in ``Architecture" means a neural architecture searched model}. \Add{We experimented with VGG16  for reference, in addition to ResNet18 and ResNet106. It should be noted that the values for ``Accuracy" are taken from the original papers, and} \Erase{In addition, }the values attached to ``Accuracy" in parentheses show the accuracy of the full-precision network described in the original paper, which has been omitted where the values are unclear.
``Time Steps" is the number of time steps related to a discrete system (\S~\ref{sec:base-forward}).
It takes a large value, which requires considerable latency during training and inference phases for typical SNNs.
Some SNN methods can have superior accuracy, whereas the latency becomes the bottleneck.
On the other hand, various BNN methods do not require multi-time step; thus, they do no latency.
However, their binarized weights deteriorate the accuracy.
Thus, there is a certain trade-off between inference accuracy and latency when pursuing computational efficiency.
Fortunately, the proposed method is superior in this respect because of its aforementioned good balance.
It is also energy-efficient.
For the Fashion-MNIST, CIFAR-100, and TinyImageNet datasets, the same comparison is summarized in Tables~\ref{tbl:comparison-fmnist-s}, \ref{tbl:comparison-cifar100-s}, and \ref{tbl:comparison-tinyimagenet-s}, respectively, where we can observe similar trends.

Fig.~\ref{fig:flops-energy} shows the FLOPs and energy consumptions of SNN models\footnote{These values are concepts for SNNs; thus, we omitted the BNNs.} for which the codes are publicly available, having more than 90\% classification accuracies, and accuracies can be reproducible in our environment. \Erase{Note that the numbers on $x$-axis correspond to those listed in Table~\ref{tbl:comparison-cifar10-s}, }\Erase{and computation details}\Add{Computational details of them} are provided in \ref{app:energy-calculation}. As shown in these figures, S$^3$NNs (\Erase{(3) and (6)}\Add{(2), (4) and (7)}) can reduce both FLOPs and energy consumption a lot from ANNs \Erase{as}\Add{the} same as \citet{chowdhury2021one} \Erase{(2)}\Add{(3)}, which realizes the single-time step SNN by iteratively reducing the time steps after training the ANN. This result indicates that we can achieve similar outputs to \citet{chowdhury2021one} without using the complex procedure by using our approach.
\Add{Note that the other approaches in Fig.~\ref{fig:flops-energy} cannot improve the energy efficiency much, which suffers from multi-time steps.}

\section{Conclusion}
\label{sec:conclusion}

In this study, we proposed the SNN with time steps $\to 1$ that is superior to the multi-time step SNNs in training and inference cost.
First, we realized this idea based on TSSL-BP~\citep{zhang2020temporal} and derived the novel surrogate gradient function for the single-step spiking neuron.
Then, we constructed S$^3$NN using the derived single-step spiking neuron and demonstrated that S$^3$NN could achieve inference accuracy comparable to the full-precision network on static image classification tasks.
We also showed that S$^3$NN had a high energy efficiency. 

Finally, we provide our limitations and future work.
First, the pros and cons for other tasks are unclear since we constructed S$^3$NN only for static image classification tasks.
Therefore, extending single-step spiking neurons to the multi-time steps and applying them to time-series tasks may be interesting, although it is outside the scope of this paper and, in a sense, inverted ideas.
Second, although the concept of deriving a single-step spiking neuron from existing SNNs may be universal, its implementation in this paper is limited to a specific method~\citep{zhang2020temporal}.
In addition to \citet{zhang2020temporal}, there are other training methods for SNNs that consider spike timing~\citep{9534390,shen2021backpropagation,yan2021graph,pmlr-v139-yang21n}.
Since different training methods have different reduction results, the single-step spiking neurons (including its surrogate gradient) derived from these methods differ from ours.
We expect some of these single-step spiking neurons to be free from hyperparameters, such as those inherited from \citet{zhang2020temporal}.
\Erase{In parallel with these derivations, we plan to analyze the impact of the asymmetricity of the surrogate gradient by searching for the optimal warm-up mechanism and train hyperparameters of the S$^3$NN derived from \citet{zhang2020temporal} as learnable parameters~\citep{wang2020sparsity} to automatically find the optimal values.
}
\Add{In parallel with these derivations, we will train hyperparameters of the S$^3$NN derived from \citet{zhang2020temporal} as learnable parameters~\citep{wang2020sparsity} to automatically find the optimal values.
In addition, we plan to analyze the impact of the asymmetricity of the surrogate gradients by searching for the optimal warm-up mechanism or introducing hyperparameters into existing methods.}

Although there is still room for further studies, we believe that our idea will become one of the essential tools for realizing environmentally friendly machines.

\bibliography{mybibfile.bib}

\begin{thebibliography}{71}
\expandafter\ifx\csname natexlab\endcsname\relax\def\natexlab#1{#1}\fi
\providecommand{\url}[1]{\texttt{#1}}
\providecommand{\href}[2]{#2}
\providecommand{\path}[1]{#1}
\providecommand{\DOIprefix}{doi:}
\providecommand{\ArXivprefix}{arXiv:}
\providecommand{\URLprefix}{URL: }
\providecommand{\Pubmedprefix}{pmid:}
\providecommand{\doi}[1]{\href{http://dx.doi.org/#1}{\path{#1}}}
\providecommand{\Pubmed}[1]{\href{pmid:#1}{\path{#1}}}
\providecommand{\bibinfo}[2]{#2}
\ifx\xfnm\relax \def\xfnm[#1]{\unskip,\space#1}\fi
\bibitem[{Akopyan et~al.(2015)Akopyan, Sawada, Cassidy, Alvarez-Icaza, Arthur,
  Merolla, Imam, Nakamura, Datta, Nam et~al.}]{akopyan2015truenorth}
\bibinfo{author}{Akopyan, F.}, \bibinfo{author}{Sawada, J.},
  \bibinfo{author}{Cassidy, A.}, \bibinfo{author}{Alvarez-Icaza, R.},
  \bibinfo{author}{Arthur, J.}, \bibinfo{author}{Merolla, P.},
  \bibinfo{author}{Imam, N.}, \bibinfo{author}{Nakamura, Y.},
  \bibinfo{author}{Datta, P.}, \bibinfo{author}{Nam, G.-J.} et~al.
  (\bibinfo{year}{2015}).
\newblock \bibinfo{title}{{TrueNorth}: Design and tool flow of a 65 m{W} 1
  million neuron programmable neurosynaptic chip}.
\newblock {\it \bibinfo{journal}{IEEE Transactions on Computer-Aided Design of
  Integrated Circuits and Systems}\/},  {\it \bibinfo{volume}{34}\/},
  \bibinfo{pages}{1537--1557}.
\bibitem[{Bengio et~al.(2013)Bengio, L{\'{e}}onard \&
  Courville}]{bengio2013estimating}
\bibinfo{author}{Bengio, Y.}, \bibinfo{author}{L{\'{e}}onard, N.}, \&
  \bibinfo{author}{Courville, A.} (\bibinfo{year}{2013}).
\newblock \bibinfo{title}{Estimating or propagating gradients through
  stochastic neurons for conditional computation}.
\newblock \href{http://arxiv.org/abs/1308.3432}{\tt arXiv:1308.3432}.
\bibitem[{Bohte et~al.(2002)Bohte, Kok \& La~Poutre}]{bohte2002error}
\bibinfo{author}{Bohte, S.~M.}, \bibinfo{author}{Kok, J.~N.}, \&
  \bibinfo{author}{La~Poutre, H.} (\bibinfo{year}{2002}).
\newblock \bibinfo{title}{Error-backpropagation in temporally encoded networks
  of spiking neurons}.
\newblock {\it \bibinfo{journal}{Neurocomputing}\/},  {\it
  \bibinfo{volume}{48}\/}, \bibinfo{pages}{17--37}.
\bibitem[{Chen et~al.(2021)Chen, Zhang, Ouyang, Liu, Shen \&
  Wang}]{chen2021bnn}
\bibinfo{author}{Chen, T.}, \bibinfo{author}{Zhang, Z.},
  \bibinfo{author}{Ouyang, X.}, \bibinfo{author}{Liu, Z.},
  \bibinfo{author}{Shen, Z.}, \& \bibinfo{author}{Wang, Z.}
  (\bibinfo{year}{2021}).
\newblock \bibinfo{title}{“{BNN-BN}=?”: Training binary neural networks
  without batch normalization}.
\newblock In {\it \bibinfo{booktitle}{2021 IEEE/CVF Conference on Computer
  Vision and Pattern Recognition Workshops (CVPRW)}\/} (pp.
  \bibinfo{pages}{4614--4624}).
\newblock \bibinfo{organization}{IEEE}.
\bibitem[{Cheng et~al.(2020)Cheng, Hao, Xu \& Xu}]{cheng2020lisnn}
\bibinfo{author}{Cheng, X.}, \bibinfo{author}{Hao, Y.}, \bibinfo{author}{Xu,
  J.}, \& \bibinfo{author}{Xu, B.} (\bibinfo{year}{2020}).
\newblock \bibinfo{title}{{LISNN}: Improving spiking neural networks with
  lateral interactions for robust object recognition}.
\newblock In \bibinfo{editor}{C.~Bessiere} (Ed.), {\it
  \bibinfo{booktitle}{Proceedings of the Twenty-Ninth International Joint
  Conference on Artificial Intelligence, {IJCAI-20}}\/} (pp.
  \bibinfo{pages}{1519--1525}).
\bibitem[{Chowdhury et~al.(2021)Chowdhury, Rathi \& Roy}]{chowdhury2021one}
\bibinfo{author}{Chowdhury, S.~S.}, \bibinfo{author}{Rathi, N.}, \&
  \bibinfo{author}{Roy, K.} (\bibinfo{year}{2021}).
\newblock \bibinfo{title}{One timestep is all you need: Training spiking neural
  networks with ultra low latency}.
\newblock \href{http://arxiv.org/abs/2110.05929}{\tt arXiv:2110.05929}.
\bibitem[{Courbariaux et~al.(2015)Courbariaux, Bengio \&
  David}]{courbariaux2015binaryconnect}
\bibinfo{author}{Courbariaux, M.}, \bibinfo{author}{Bengio, Y.}, \&
  \bibinfo{author}{David, J.-P.} (\bibinfo{year}{2015}).
\newblock \bibinfo{title}{{BinaryConnect}: Training deep neural networks with
  binary weights during propagations}.
\newblock {\it \bibinfo{journal}{Advances in neural information processing
  systems}\/},  {\it \bibinfo{volume}{28}\/}.
\bibitem[{Cubuk et~al.(2020)Cubuk, Zoph, Shlens \& Le}]{cubuk2020randaugment}
\bibinfo{author}{Cubuk, E.~D.}, \bibinfo{author}{Zoph, B.},
  \bibinfo{author}{Shlens, J.}, \& \bibinfo{author}{Le, Q.~V.}
  (\bibinfo{year}{2020}).
\newblock \bibinfo{title}{{Randaugment}: Practical automated data augmentation
  with a reduced search space}.
\newblock In {\it \bibinfo{booktitle}{Proceedings of the IEEE/CVF Conference on
  Computer Vision and Pattern Recognition Workshops}\/} (pp.
  \bibinfo{pages}{702--703}).
\bibitem[{Davies et~al.(2018)Davies, Srinivasa, Lin, Chinya, Cao, Choday,
  Dimou, Joshi, Imam, Jain et~al.}]{davies2018loihi}
\bibinfo{author}{Davies, M.}, \bibinfo{author}{Srinivasa, N.},
  \bibinfo{author}{Lin, T.-H.}, \bibinfo{author}{Chinya, G.},
  \bibinfo{author}{Cao, Y.}, \bibinfo{author}{Choday, S.~H.},
  \bibinfo{author}{Dimou, G.}, \bibinfo{author}{Joshi, P.},
  \bibinfo{author}{Imam, N.}, \bibinfo{author}{Jain, S.} et~al.
  (\bibinfo{year}{2018}).
\newblock \bibinfo{title}{Loihi: A neuromorphic manycore processor with on-chip
  learning}.
\newblock {\it \bibinfo{journal}{IEEE Micro}\/},  {\it \bibinfo{volume}{38}\/},
  \bibinfo{pages}{82--99}.
\bibitem[{Deng \& Gu(2021)}]{deng2020optimal}
\bibinfo{author}{Deng, S.}, \& \bibinfo{author}{Gu, S.} (\bibinfo{year}{2021}).
\newblock \bibinfo{title}{Optimal conversion of conventional artificial neural
  networks to spiking neural networks}.
\newblock In {\it \bibinfo{booktitle}{International Conference on Learning
  Representations}\/}.
\bibitem[{Diehl et~al.(2015)Diehl, Neil, Binas, Cook, Liu \&
  Pfeiffer}]{diehl2015fast}
\bibinfo{author}{Diehl, P.~U.}, \bibinfo{author}{Neil, D.},
  \bibinfo{author}{Binas, J.}, \bibinfo{author}{Cook, M.},
  \bibinfo{author}{Liu, S.-C.}, \& \bibinfo{author}{Pfeiffer, M.}
  (\bibinfo{year}{2015}).
\newblock \bibinfo{title}{Fast-classifying, high-accuracy spiking deep networks
  through weight and threshold balancing}.
\newblock In {\it \bibinfo{booktitle}{2015 International joint conference on
  neural networks (IJCNN)}\/} (pp. \bibinfo{pages}{1--8}).
\bibitem[{Ding et~al.(2021)Ding, Yu, Tian \& Huang}]{ding2021optimal}
\bibinfo{author}{Ding, J.}, \bibinfo{author}{Yu, Z.}, \bibinfo{author}{Tian,
  Y.}, \& \bibinfo{author}{Huang, T.} (\bibinfo{year}{2021}).
\newblock \bibinfo{title}{Optimal {ANN-SNN} conversion for fast and accurate
  inference in deep spiking neural networks}.
\newblock In \bibinfo{editor}{Z.-H. Zhou} (Ed.), {\it
  \bibinfo{booktitle}{Proceedings of the Thirtieth International Joint
  Conference on Artificial Intelligence, {IJCAI-21}}\/} (pp.
  \bibinfo{pages}{2328--2336}).
\newblock \bibinfo{publisher}{International Joint Conferences on Artificial
  Intelligence Organization}.
\bibitem[{Esser et~al.(2016)Esser, Merolla, Arthur, Cassidy, Appuswamy,
  Andreopoulos, Berg, McKinstry, Melano, Barch et~al.}]{esser2016cover}
\bibinfo{author}{Esser, S.~K.}, \bibinfo{author}{Merolla, P.~A.},
  \bibinfo{author}{Arthur, J.~V.}, \bibinfo{author}{Cassidy, A.~S.},
  \bibinfo{author}{Appuswamy, R.}, \bibinfo{author}{Andreopoulos, A.},
  \bibinfo{author}{Berg, D.~J.}, \bibinfo{author}{McKinstry, J.~L.},
  \bibinfo{author}{Melano, T.}, \bibinfo{author}{Barch, D.~R.} et~al.
  (\bibinfo{year}{2016}).
\newblock \bibinfo{title}{Convolutional networks for fast, energy-efficient
  neuromorphic computing}.
\newblock {\it \bibinfo{journal}{Proceedings of the National Academy of
  Sciences of the United States of America}\/},  {\it \bibinfo{volume}{113}\/},
  \bibinfo{pages}{11441}.
\bibitem[{Fang et~al.(2021)Fang, Yu, Chen, Masquelier, Huang \&
  Tian}]{fang2020incorporating}
\bibinfo{author}{Fang, W.}, \bibinfo{author}{Yu, Z.}, \bibinfo{author}{Chen,
  Y.}, \bibinfo{author}{Masquelier, T.}, \bibinfo{author}{Huang, T.}, \&
  \bibinfo{author}{Tian, Y.} (\bibinfo{year}{2021}).
\newblock \bibinfo{title}{Incorporating learnable membrane time constant to
  enhance learning of spiking neural networks}.
\newblock In {\it \bibinfo{booktitle}{Proceedings of the IEEE/CVF International
  Conference on Computer Vision}\/} (pp. \bibinfo{pages}{2661--2671}).
\bibitem[{Han et~al.(2020)Han, Srinivasan \& Roy}]{han2020rmp}
\bibinfo{author}{Han, B.}, \bibinfo{author}{Srinivasan, G.}, \&
  \bibinfo{author}{Roy, K.} (\bibinfo{year}{2020}).
\newblock \bibinfo{title}{{RMP-SNN}: Residual membrane potential neuron for
  enabling deeper high-accuracy and low-latency spiking neural network}.
\newblock In {\it \bibinfo{booktitle}{Proceedings of the IEEE/CVF Conference on
  Computer Vision and Pattern Recognition}\/} (pp.
  \bibinfo{pages}{13558--13567}).
\bibitem[{He et~al.(2016)He, Zhang, Ren \& Sun}]{he2016identity}
\bibinfo{author}{He, K.}, \bibinfo{author}{Zhang, X.}, \bibinfo{author}{Ren,
  S.}, \& \bibinfo{author}{Sun, J.} (\bibinfo{year}{2016}).
\newblock \bibinfo{title}{Identity mappings in deep residual networks}.
\newblock In {\it \bibinfo{booktitle}{Computer Vision--ECCV 2016: 14th European
  Conference, Amsterdam, The Netherlands, October 11--14, 2016, Proceedings,
  Part IV 14}\/} (pp. \bibinfo{pages}{630--645}).
\newblock \bibinfo{organization}{Springer}.
\bibitem[{Hinton et~al.(2006)Hinton, Osindero \& Teh}]{hinton2006fast}
\bibinfo{author}{Hinton, G.~E.}, \bibinfo{author}{Osindero, S.}, \&
  \bibinfo{author}{Teh, Y.-W.} (\bibinfo{year}{2006}).
\newblock \bibinfo{title}{A fast learning algorithm for deep belief nets}.
\newblock {\it \bibinfo{journal}{Neural Computation}\/},  {\it
  \bibinfo{volume}{18}\/}, \bibinfo{pages}{1527--1554}.
\bibitem[{Horowitz(2014)}]{horowitz20141}
\bibinfo{author}{Horowitz, M.} (\bibinfo{year}{2014}).
\newblock \bibinfo{title}{1.1 {C}omputing's energy problem (and what we can do
  about it)}.
\newblock In {\it \bibinfo{booktitle}{2014 IEEE International Solid-State
  Circuits Conference Digest of Technical Papers (ISSCC)}\/} (pp.
  \bibinfo{pages}{10--14}).
\bibitem[{Hubara et~al.(2016)Hubara, Courbariaux, Soudry, El-Yaniv \&
  Bengio}]{hubara2016binarized}
\bibinfo{author}{Hubara, I.}, \bibinfo{author}{Courbariaux, M.},
  \bibinfo{author}{Soudry, D.}, \bibinfo{author}{El-Yaniv, R.}, \&
  \bibinfo{author}{Bengio, Y.} (\bibinfo{year}{2016}).
\newblock \bibinfo{title}{Binarized neural networks}.
\newblock {\it \bibinfo{journal}{Advances in Neural Information Processing
  Systems}\/},  {\it \bibinfo{volume}{29}\/}.
\bibitem[{Kheradpisheh et~al.(2021)Kheradpisheh, Mirsadeghi \&
  Masquelier}]{kheradpisheh2020bs4nn}
\bibinfo{author}{Kheradpisheh, S.~R.}, \bibinfo{author}{Mirsadeghi, M.}, \&
  \bibinfo{author}{Masquelier, T.} (\bibinfo{year}{2021}).
\newblock \bibinfo{title}{{BS4NN}: Binarized spiking neural networks with
  temporal coding and learning}.
\newblock {\it \bibinfo{journal}{Neural Processing Letters}\/},  {\it
  \bibinfo{volume}{53}\/}, \bibinfo{pages}{1--19}.
\bibitem[{Kim et~al.(2022{\natexlab{a}})Kim, Li, Park, Venkatesha \&
  Panda}]{kim2022neural}
\bibinfo{author}{Kim, Y.}, \bibinfo{author}{Li, Y.}, \bibinfo{author}{Park,
  H.}, \bibinfo{author}{Venkatesha, Y.}, \& \bibinfo{author}{Panda, P.}
  (\bibinfo{year}{2022}{\natexlab{a}}).
\newblock \bibinfo{title}{Neural architecture search for spiking neural
  networks}.
\newblock In {\it \bibinfo{booktitle}{Computer Vision--ECCV 2022: 17th European
  Conference, Tel Aviv, Israel, October 23--27, 2022, Proceedings, Part
  XXIV}\/} (pp. \bibinfo{pages}{36--56}).
\newblock \bibinfo{organization}{Springer}.
\bibitem[{Kim \& Panda(2021)}]{kim2020revisiting}
\bibinfo{author}{Kim, Y.}, \& \bibinfo{author}{Panda, P.}
  (\bibinfo{year}{2021}).
\newblock \bibinfo{title}{Revisiting batch normalization for training
  low-latency deep spiking neural networks from scratch}.
\newblock {\it \bibinfo{journal}{Frontiers in Neuroscience}\/},  {\it
  \bibinfo{volume}{15}\/}.
\bibitem[{Kim et~al.(2022{\natexlab{b}})Kim, Venkatesha \&
  Panda}]{kim2022privatesnn}
\bibinfo{author}{Kim, Y.}, \bibinfo{author}{Venkatesha, Y.}, \&
  \bibinfo{author}{Panda, P.} (\bibinfo{year}{2022}{\natexlab{b}}).
\newblock \bibinfo{title}{Private{SNN}: Privacy-preserving spiking neural
  networks}.
\newblock In {\it \bibinfo{booktitle}{Proceedings of the AAAI Conference on
  Artificial Intelligence}\/} (pp. \bibinfo{pages}{1192--1200}).
\newblock volume~\bibinfo{volume}{36}.
\bibitem[{Krizhevsky(2009)}]{krizhevsky2009learning}
\bibinfo{author}{Krizhevsky, A.} (\bibinfo{year}{2009}).
\newblock {\it \bibinfo{title}{Learning multiple layers of features from tiny
  images}\/}.
\newblock \bibinfo{type}{Technical Report}.
\bibitem[{Krizhevsky et~al.(2012)Krizhevsky, Sutskever \&
  Hinton}]{krizhevsky2012imagenet}
\bibinfo{author}{Krizhevsky, A.}, \bibinfo{author}{Sutskever, I.}, \&
  \bibinfo{author}{Hinton, G.~E.} (\bibinfo{year}{2012}).
\newblock \bibinfo{title}{{ImageNet} classification with deep convolutional
  neural networks}.
\newblock {\it \bibinfo{journal}{Advances in Neural Information Processing
  Systems}\/},  {\it \bibinfo{volume}{25}\/}, \bibinfo{pages}{1097--1105}.
\bibitem[{Kundu et~al.(2021)Kundu, Datta, Pedram \& Beerel}]{kundu2021towards}
\bibinfo{author}{Kundu, S.}, \bibinfo{author}{Datta, G.},
  \bibinfo{author}{Pedram, M.}, \& \bibinfo{author}{Beerel, P.~A.}
  (\bibinfo{year}{2021}).
\newblock \bibinfo{title}{Towards low-latency energy-efficient deep {SNN}s via
  attention-guided compression}.
\newblock \href{http://arxiv.org/abs/2107.12445}{\tt arXiv:2107.12445}.
\bibitem[{Le \& Yang(2015)}]{Le2015TinyIV}
\bibinfo{author}{Le, Y.}, \& \bibinfo{author}{Yang, X.~S.}
  (\bibinfo{year}{2015}).
\newblock \bibinfo{title}{Tiny {ImageNet} visual recognition challenge}.
\bibitem[{Lee et~al.(2020)Lee, Sarwar, Panda, Srinivasan \&
  Roy}]{lee2020enabling}
\bibinfo{author}{Lee, C.}, \bibinfo{author}{Sarwar, S.~S.},
  \bibinfo{author}{Panda, P.}, \bibinfo{author}{Srinivasan, G.}, \&
  \bibinfo{author}{Roy, K.} (\bibinfo{year}{2020}).
\newblock \bibinfo{title}{Enabling spike-based backpropagation for training
  deep neural network architectures}.
\newblock {\it \bibinfo{journal}{Frontiers in Neuroscience}\/},  {\it
  \bibinfo{volume}{14}\/}, \bibinfo{pages}{119}.
\bibitem[{Li et~al.(2021)Li, Deng, Dong, Gong \& Gu}]{pmlr-v139-li21d}
\bibinfo{author}{Li, Y.}, \bibinfo{author}{Deng, S.}, \bibinfo{author}{Dong,
  X.}, \bibinfo{author}{Gong, R.}, \& \bibinfo{author}{Gu, S.}
  (\bibinfo{year}{2021}).
\newblock \bibinfo{title}{A free lunch from {ANN}: Towards efficient, accurate
  spiking neural networks calibration}.
\newblock In {\it \bibinfo{booktitle}{International Conference on Machine
  Learning}\/} (pp. \bibinfo{pages}{6316--6325}).
\newblock \bibinfo{organization}{PMLR}.
\bibitem[{Li et~al.(2022)Li, Zhao \& Zeng}]{li2021bsnn}
\bibinfo{author}{Li, Y.}, \bibinfo{author}{Zhao, D.}, \& \bibinfo{author}{Zeng,
  Y.} (\bibinfo{year}{2022}).
\newblock \bibinfo{title}{{BSNN}: Towards faster and better conversion of
  artificial neural networks to spiking neural networks with bistable neurons}.
\newblock {\it \bibinfo{journal}{Frontiers in neuroscience}\/},  {\it
  \bibinfo{volume}{16}\/}, \bibinfo{pages}{991851}.
\bibitem[{Loshchilov \& Hutter(2017)}]{loshchilov2016sgdr}
\bibinfo{author}{Loshchilov, I.}, \& \bibinfo{author}{Hutter, F.}
  (\bibinfo{year}{2017}).
\newblock \bibinfo{title}{{SGDR}: Stochastic gradient descent with warm
  restarts}.
\newblock In {\it \bibinfo{booktitle}{International Conference on Learning
  Representations}\/}.
\bibitem[{Lu \& Sengupta(2020)}]{lu2020exploring}
\bibinfo{author}{Lu, S.}, \& \bibinfo{author}{Sengupta, A.}
  (\bibinfo{year}{2020}).
\newblock \bibinfo{title}{Exploring the connection between binary and spiking
  neural networks}.
\newblock {\it \bibinfo{journal}{Frontiers in Neuroscience}\/},  {\it
  \bibinfo{volume}{14}\/}, \bibinfo{pages}{535}.
\bibitem[{Ma et~al.(2021)Ma, Xu \& Yu}]{9534390}
\bibinfo{author}{Ma, C.}, \bibinfo{author}{Xu, J.}, \& \bibinfo{author}{Yu, Q.}
  (\bibinfo{year}{2021}).
\newblock \bibinfo{title}{Temporal dependent local learning for deep spiking
  neural networks}.
\newblock In {\it \bibinfo{booktitle}{2021 International Joint Conference on
  Neural Networks (IJCNN)}\/} (pp. \bibinfo{pages}{1--7}).
\bibitem[{Maass(1997)}]{maass1997networks}
\bibinfo{author}{Maass, W.} (\bibinfo{year}{1997}).
\newblock \bibinfo{title}{Networks of spiking neurons: The third generation of
  neural network models}.
\newblock {\it \bibinfo{journal}{Neural Networks}\/},  {\it
  \bibinfo{volume}{10}\/}, \bibinfo{pages}{1659--1671}.
\bibitem[{Maguire et~al.(2007)Maguire, McGinnity, Glackin, Ghani, Belatreche \&
  Harkin}]{maguire2007challenges}
\bibinfo{author}{Maguire, L.~P.}, \bibinfo{author}{McGinnity, T.~M.},
  \bibinfo{author}{Glackin, B.}, \bibinfo{author}{Ghani, A.},
  \bibinfo{author}{Belatreche, A.}, \& \bibinfo{author}{Harkin, J.}
  (\bibinfo{year}{2007}).
\newblock \bibinfo{title}{Challenges for large-scale implementations of spiking
  neural networks on {FPGA}s}.
\newblock {\it \bibinfo{journal}{Neurocomputing}\/},  {\it
  \bibinfo{volume}{71}\/}, \bibinfo{pages}{13--29}.
\bibitem[{McCulloch \& Pitts(1943)}]{mcculloch1943logical}
\bibinfo{author}{McCulloch, W.~S.}, \& \bibinfo{author}{Pitts, W.}
  (\bibinfo{year}{1943}).
\newblock \bibinfo{title}{A logical calculus of the ideas immanent in nervous
  activity}.
\newblock {\it \bibinfo{journal}{The Bulletin of Mathematical Biophysics}\/},
  {\it \bibinfo{volume}{5}\/}, \bibinfo{pages}{115--133}.
\bibitem[{Misra \& Saha(2010)}]{misra2010artificial}
\bibinfo{author}{Misra, J.}, \& \bibinfo{author}{Saha, I.}
  (\bibinfo{year}{2010}).
\newblock \bibinfo{title}{Artificial neural networks in hardware: A survey of
  two decades of progress}.
\newblock {\it \bibinfo{journal}{Neurocomputing}\/},  {\it
  \bibinfo{volume}{74}\/}, \bibinfo{pages}{239--255}.
\bibitem[{Na et~al.(2022)Na, Mok, Park, Lee, Choe \& Yoon}]{na2022autosnn}
\bibinfo{author}{Na, B.}, \bibinfo{author}{Mok, J.}, \bibinfo{author}{Park,
  S.}, \bibinfo{author}{Lee, D.}, \bibinfo{author}{Choe, H.}, \&
  \bibinfo{author}{Yoon, S.} (\bibinfo{year}{2022}).
\newblock \bibinfo{title}{{AutoSNN}: Towards energy-efficient spiking neural
  networks}.
\newblock In {\it \bibinfo{booktitle}{International Conference on Machine
  Learning}\/} (pp. \bibinfo{pages}{16253--16269}).
\newblock \bibinfo{organization}{PMLR}.
\bibitem[{Neftci et~al.(2019)Neftci, Mostafa \& Zenke}]{neftci2019surrogate}
\bibinfo{author}{Neftci, E.~O.}, \bibinfo{author}{Mostafa, H.}, \&
  \bibinfo{author}{Zenke, F.} (\bibinfo{year}{2019}).
\newblock \bibinfo{title}{Surrogate gradient learning in spiking neural
  networks: Bringing the power of gradient-based optimization to spiking neural
  networks}.
\newblock {\it \bibinfo{journal}{IEEE Signal Processing Magazine}\/},  {\it
  \bibinfo{volume}{36}\/}, \bibinfo{pages}{51--63}.
\bibitem[{Neil et~al.(2016)Neil, Pfeiffer \& Liu}]{neil2016learning}
\bibinfo{author}{Neil, D.}, \bibinfo{author}{Pfeiffer, M.}, \&
  \bibinfo{author}{Liu, S.-C.} (\bibinfo{year}{2016}).
\newblock \bibinfo{title}{Learning to be efficient: Algorithms for training
  low-latency, low-compute deep spiking neural networks}.
\newblock In {\it \bibinfo{booktitle}{Proceedings of the 31st annual ACM
  symposium on applied computing}\/} (pp. \bibinfo{pages}{293--298}).
\bibitem[{Qiao et~al.(2020)Qiao, Hu, Chen, Rong, Ning, Yu \&
  Liu}]{qiao2020stbnn}
\bibinfo{author}{Qiao, G.}, \bibinfo{author}{Hu, S.}, \bibinfo{author}{Chen,
  T.}, \bibinfo{author}{Rong, L.}, \bibinfo{author}{Ning, N.},
  \bibinfo{author}{Yu, Q.}, \& \bibinfo{author}{Liu, Y.}
  (\bibinfo{year}{2020}).
\newblock \bibinfo{title}{{STBNN}: Hardware-friendly spatio-temporal binary
  neural network with high pattern recognition accuracy}.
\newblock {\it \bibinfo{journal}{Neurocomputing}\/},  {\it
  \bibinfo{volume}{409}\/}, \bibinfo{pages}{351--360}.
\bibitem[{Rastegari et~al.(2016)Rastegari, Ordonez, Redmon \&
  Farhadi}]{rastegari2016xnor}
\bibinfo{author}{Rastegari, M.}, \bibinfo{author}{Ordonez, V.},
  \bibinfo{author}{Redmon, J.}, \& \bibinfo{author}{Farhadi, A.}
  (\bibinfo{year}{2016}).
\newblock \bibinfo{title}{{XNOR-Net}: {ImageNet} classification using binary
  convolutional neural networks}.
\newblock In {\it \bibinfo{booktitle}{Computer Vision--ECCV 2016: 14th European
  Conference, Amsterdam, The Netherlands, October 11--14, 2016, Proceedings,
  Part IV}\/} (pp. \bibinfo{pages}{525--542}).
\newblock \bibinfo{organization}{Springer}.
\bibitem[{Rathi \& Roy(2020)}]{rathi2020diet}
\bibinfo{author}{Rathi, N.}, \& \bibinfo{author}{Roy, K.}
  (\bibinfo{year}{2020}).
\newblock \bibinfo{title}{{DIET-SNN}: Direct input encoding with leakage and
  threshold optimization in deep spiking neural networks}.
\newblock \href{http://arxiv.org/abs/2008.03658}{\tt arXiv:2008.03658}.
\bibitem[{Rathi et~al.(2019)Rathi, Srinivasan, Panda \&
  Roy}]{rathi2019enabling}
\bibinfo{author}{Rathi, N.}, \bibinfo{author}{Srinivasan, G.},
  \bibinfo{author}{Panda, P.}, \& \bibinfo{author}{Roy, K.}
  (\bibinfo{year}{2019}).
\newblock \bibinfo{title}{Enabling deep spiking neural networks with hybrid
  conversion and spike timing dependent backpropagation}.
\newblock In {\it \bibinfo{booktitle}{International Conference on Learning
  Representations}\/}.
\bibitem[{van Rossum(2001)}]{van2001novel}
\bibinfo{author}{van Rossum, M.~C.} (\bibinfo{year}{2001}).
\newblock \bibinfo{title}{A novel spike distance}.
\newblock {\it \bibinfo{journal}{Neural Computation}\/},  {\it
  \bibinfo{volume}{13}\/}, \bibinfo{pages}{751--763}.
\bibitem[{Roy et~al.(2019)Roy, Chakraborty \& Roy}]{roy2019scaling}
\bibinfo{author}{Roy, D.}, \bibinfo{author}{Chakraborty, I.}, \&
  \bibinfo{author}{Roy, K.} (\bibinfo{year}{2019}).
\newblock \bibinfo{title}{Scaling deep spiking neural networks with binary
  stochastic activations}.
\newblock In {\it \bibinfo{booktitle}{2019 IEEE International Conference on
  Cognitive Computing (ICCC)}\/} (pp. \bibinfo{pages}{50--58}).
\bibitem[{Rueckauer et~al.(2017)Rueckauer, Lungu, Hu, Pfeiffer \&
  Liu}]{rueckauer2017conversion}
\bibinfo{author}{Rueckauer, B.}, \bibinfo{author}{Lungu, I.-A.},
  \bibinfo{author}{Hu, Y.}, \bibinfo{author}{Pfeiffer, M.}, \&
  \bibinfo{author}{Liu, S.-C.} (\bibinfo{year}{2017}).
\newblock \bibinfo{title}{Conversion of continuous-valued deep networks to
  efficient event-driven networks for image classification}.
\newblock {\it \bibinfo{journal}{Frontiers in Neuroscience}\/},  {\it
  \bibinfo{volume}{11}\/}, \bibinfo{pages}{682}.
\bibitem[{Salakhutdinov \& Hinton(2009)}]{salakhutdinov2009deep}
\bibinfo{author}{Salakhutdinov, R.}, \& \bibinfo{author}{Hinton, G.}
  (\bibinfo{year}{2009}).
\newblock \bibinfo{title}{Deep {B}oltzmann machines}.
\newblock In {\it \bibinfo{booktitle}{Proceedings of the Twelth International
  Conference on Artificial Intelligence and Statistics}\/} (pp.
  \bibinfo{pages}{448--455}).
\bibitem[{Severa et~al.(2019)Severa, Vineyard, Dellana, Verzi \&
  Aimone}]{severa2018whetstone}
\bibinfo{author}{Severa, W.}, \bibinfo{author}{Vineyard, C.~M.},
  \bibinfo{author}{Dellana, R.}, \bibinfo{author}{Verzi, S.~J.}, \&
  \bibinfo{author}{Aimone, J.~B.} (\bibinfo{year}{2019}).
\newblock \bibinfo{title}{Training deep neural networks for binary
  communication with the whetstone method}.
\newblock {\it \bibinfo{journal}{Nature Machine Intelligence}\/},  {\it
  \bibinfo{volume}{1}\/}, \bibinfo{pages}{86--94}.
\bibitem[{Shekhovtsov \& Yanush(2021)}]{shekhovtsov2021reintroducing}
\bibinfo{author}{Shekhovtsov, A.}, \& \bibinfo{author}{Yanush, V.}
  (\bibinfo{year}{2021}).
\newblock \bibinfo{title}{Reintroducing straight-through estimators as
  principled methods for stochastic binary networks}.
\newblock In {\it \bibinfo{booktitle}{DAGM German Conference on Pattern
  Recognition}\/} (pp. \bibinfo{pages}{111--126}).
\bibitem[{Shen et~al.(2022)Shen, Zhao \& Zeng}]{shen2021backpropagation}
\bibinfo{author}{Shen, G.}, \bibinfo{author}{Zhao, D.}, \&
  \bibinfo{author}{Zeng, Y.} (\bibinfo{year}{2022}).
\newblock \bibinfo{title}{Backpropagation with biologically plausible
  spatiotemporal adjustment for training deep spiking neural networks}.
\newblock {\it \bibinfo{journal}{Patterns}\/},  {\it \bibinfo{volume}{3}\/},
  \bibinfo{pages}{100522}.
\bibitem[{Shen et~al.(2020)Shen, Liu, Gong \& Han}]{shen2020balanced}
\bibinfo{author}{Shen, M.}, \bibinfo{author}{Liu, X.}, \bibinfo{author}{Gong,
  R.}, \& \bibinfo{author}{Han, K.} (\bibinfo{year}{2020}).
\newblock \bibinfo{title}{Balanced binary neural networks with gated residual}.
\newblock In {\it \bibinfo{booktitle}{ICASSP 2020-2020 IEEE International
  Conference on Acoustics, Speech and Signal Processing (ICASSP)}\/} (pp.
  \bibinfo{pages}{4197--4201}).
\bibitem[{Shi et~al.(2020)Shi, Li, Zhang, Wu \& Ren}]{shi2020accurate}
\bibinfo{author}{Shi, Y.}, \bibinfo{author}{Li, H.}, \bibinfo{author}{Zhang,
  H.}, \bibinfo{author}{Wu, Z.}, \& \bibinfo{author}{Ren, S.}
  (\bibinfo{year}{2020}).
\newblock \bibinfo{title}{Accurate and efficient {LIF}-nets for 3d detection
  and recognition}.
\newblock {\it \bibinfo{journal}{IEEE Access}\/},  {\it \bibinfo{volume}{8}\/},
  \bibinfo{pages}{98562--98571}.
\bibitem[{Shrestha \& Orchard(2018)}]{shrestha2018slayer}
\bibinfo{author}{Shrestha, S.~B.}, \& \bibinfo{author}{Orchard, G.}
  (\bibinfo{year}{2018}).
\newblock \bibinfo{title}{{SLAYER}: Spike layer error reassignment in time}.
\newblock {\it \bibinfo{journal}{Advances in neural information processing
  systems}\/},  {\it \bibinfo{volume}{31}\/}.
\bibitem[{Srivastava et~al.(2014)Srivastava, Hinton, Krizhevsky, Sutskever \&
  Salakhutdinov}]{srivastava2014dropout}
\bibinfo{author}{Srivastava, N.}, \bibinfo{author}{Hinton, G.},
  \bibinfo{author}{Krizhevsky, A.}, \bibinfo{author}{Sutskever, I.}, \&
  \bibinfo{author}{Salakhutdinov, R.} (\bibinfo{year}{2014}).
\newblock \bibinfo{title}{Dropout: A simple way to prevent neural networks from
  overfitting}.
\newblock {\it \bibinfo{journal}{Journal of Machine Learning Research}\/},
  {\it \bibinfo{volume}{15}\/}, \bibinfo{pages}{1929--1958}.
\bibitem[{Szegedy et~al.(2016)Szegedy, Vanhoucke, Ioffe, Shlens \&
  Wojna}]{szegedy2016rethinking}
\bibinfo{author}{Szegedy, C.}, \bibinfo{author}{Vanhoucke, V.},
  \bibinfo{author}{Ioffe, S.}, \bibinfo{author}{Shlens, J.}, \&
  \bibinfo{author}{Wojna, Z.} (\bibinfo{year}{2016}).
\newblock \bibinfo{title}{Rethinking the inception architecture for computer
  vision}.
\newblock In {\it \bibinfo{booktitle}{Proceedings of the IEEE Conference on
  Computer Vision and Pattern Recognition (CVPR)}\/} (pp.
  \bibinfo{pages}{2818--2826}).
\bibitem[{Wang et~al.(2020{\natexlab{a}})Wang, He, Li, Zhao \&
  Cheng}]{wang2020sparsity}
\bibinfo{author}{Wang, P.}, \bibinfo{author}{He, X.}, \bibinfo{author}{Li, G.},
  \bibinfo{author}{Zhao, T.}, \& \bibinfo{author}{Cheng, J.}
  (\bibinfo{year}{2020}{\natexlab{a}}).
\newblock \bibinfo{title}{Sparsity-inducing binarized neural networks}.
\newblock In {\it \bibinfo{booktitle}{Proceedings of the AAAI Conference on
  Artificial Intelligence}\/} (pp. \bibinfo{pages}{12192--12199}).
\newblock volume~\bibinfo{volume}{34}.
\bibitem[{Wang et~al.(2020{\natexlab{b}})Wang, Lin \&
  Dang}]{wang2020supervised}
\bibinfo{author}{Wang, X.}, \bibinfo{author}{Lin, X.}, \&
  \bibinfo{author}{Dang, X.} (\bibinfo{year}{2020}{\natexlab{b}}).
\newblock \bibinfo{title}{Supervised learning in spiking neural networks: A
  review of algorithms and evaluations}.
\newblock {\it \bibinfo{journal}{Neural Networks}\/},  {\it
  \bibinfo{volume}{125}\/}, \bibinfo{pages}{258--280}.
\bibitem[{Wang et~al.(2019)Wang, Lu, Tao, Zhou \& Tian}]{wang2019learning}
\bibinfo{author}{Wang, Z.}, \bibinfo{author}{Lu, J.}, \bibinfo{author}{Tao,
  C.}, \bibinfo{author}{Zhou, J.}, \& \bibinfo{author}{Tian, Q.}
  (\bibinfo{year}{2019}).
\newblock \bibinfo{title}{Learning channel-wise interactions for binary
  convolutional neural networks}.
\newblock In {\it \bibinfo{booktitle}{Proceedings of the IEEE/CVF Conference on
  Computer Vision and Pattern Recognition}\/} (pp. \bibinfo{pages}{568--577}).
\bibitem[{Werbos(1990)}]{werbos1990backpropagation}
\bibinfo{author}{Werbos, P.~J.} (\bibinfo{year}{1990}).
\newblock \bibinfo{title}{Backpropagation through time: What it does and how to
  do it}.
\newblock {\it \bibinfo{journal}{Proceedings of the IEEE}\/},  {\it
  \bibinfo{volume}{78}\/}, \bibinfo{pages}{1550--1560}.
\bibitem[{Xiao et~al.(2017)Xiao, Rasul \& Vollgraf}]{xiao2017fashion}
\bibinfo{author}{Xiao, H.}, \bibinfo{author}{Rasul, K.}, \&
  \bibinfo{author}{Vollgraf, R.} (\bibinfo{year}{2017}).
\newblock \bibinfo{title}{Fashion-{MNIST}: A novel image dataset for
  benchmarking machine learning algorithms}.
\newblock \URLprefix \url{https://github.com/zalandoresearch/fashion-mnist}.
  \href{http://arxiv.org/abs/1708.07747}{\tt arXiv:1708.07747}.
\bibitem[{Xu \& Cheung(2019)}]{xu2019accurate}
\bibinfo{author}{Xu, Z.}, \& \bibinfo{author}{Cheung, R.~C.}
  (\bibinfo{year}{2019}).
\newblock \bibinfo{title}{Accurate and compact convolutional neural networks
  with trained binarization}.
\newblock In {\it \bibinfo{booktitle}{30th British Machine Vision Conference
  (BMVC 2019)}\/}.
\bibitem[{Yan et~al.(2021{\natexlab{a}})Yan, Chu, Chen, Jin, Huan, Zheng \&
  Zou}]{yan2021graph}
\bibinfo{author}{Yan, Y.}, \bibinfo{author}{Chu, H.}, \bibinfo{author}{Chen,
  X.}, \bibinfo{author}{Jin, Y.}, \bibinfo{author}{Huan, Y.},
  \bibinfo{author}{Zheng, L.}, \& \bibinfo{author}{Zou, Z.}
  (\bibinfo{year}{2021}{\natexlab{a}}).
\newblock \bibinfo{title}{Graph-based spatio-temporal backpropagation for
  training spiking neural networks}.
\newblock In {\it \bibinfo{booktitle}{2021 IEEE 3rd International Conference on
  Artificial Intelligence Circuits and Systems (AICAS)}\/} (pp.
  \bibinfo{pages}{1--4}).
\bibitem[{Yan et~al.(2021{\natexlab{b}})Yan, Zhou \& Wong}]{yan2021near}
\bibinfo{author}{Yan, Z.}, \bibinfo{author}{Zhou, J.}, \&
  \bibinfo{author}{Wong, W.-F.} (\bibinfo{year}{2021}{\natexlab{b}}).
\newblock \bibinfo{title}{Near lossless transfer learning for spiking neural
  networks}.
\newblock {\it \bibinfo{journal}{Proceedings of the AAAI Conference on
  Artificial Intelligence}\/},  {\it \bibinfo{volume}{35}\/},
  \bibinfo{pages}{10577--10584}.
\bibitem[{Yang et~al.(2021)Yang, Zhang \& Li}]{pmlr-v139-yang21n}
\bibinfo{author}{Yang, Y.}, \bibinfo{author}{Zhang, W.}, \&
  \bibinfo{author}{Li, P.} (\bibinfo{year}{2021}).
\newblock \bibinfo{title}{Backpropagated neighborhood aggregation for accurate
  training of spiking neural networks}.
\newblock In \bibinfo{editor}{M.~Meila}, \& \bibinfo{editor}{T.~Zhang} (Eds.),
  {\it \bibinfo{booktitle}{Proceedings of the 38th International Conference on
  Machine Learning}\/} (pp. \bibinfo{pages}{11852--11862}).
\newblock volume \bibinfo{volume}{139}.
\bibitem[{Yin et~al.(2019)Yin, Lyu, Zhang, Osher, Qi \&
  Xin}]{yin2019understanding}
\bibinfo{author}{Yin, P.}, \bibinfo{author}{Lyu, J.}, \bibinfo{author}{Zhang,
  S.}, \bibinfo{author}{Osher, S.}, \bibinfo{author}{Qi, Y.}, \&
  \bibinfo{author}{Xin, J.} (\bibinfo{year}{2019}).
\newblock \bibinfo{title}{Understanding straight-through estimator in training
  activation quantized neural nets}.
\newblock In {\it \bibinfo{booktitle}{International Conference on Learning
  Representations}\/}.
\bibitem[{Yuan \& Agaian(2021)}]{yuan2021comprehensive}
\bibinfo{author}{Yuan, C.}, \& \bibinfo{author}{Agaian, S.~S.}
  (\bibinfo{year}{2021}).
\newblock \bibinfo{title}{A comprehensive review of binary neural network}.
\newblock \href{http://arxiv.org/abs/2110.06804}{\tt arXiv:2110.06804}.
\bibitem[{Zenke \& Ganguli(2018)}]{zenke2018superspike}
\bibinfo{author}{Zenke, F.}, \& \bibinfo{author}{Ganguli, S.}
  (\bibinfo{year}{2018}).
\newblock \bibinfo{title}{{SuperSpike}: Supervised learning in multilayer
  spiking neural networks}.
\newblock {\it \bibinfo{journal}{Neural Computation}\/},  {\it
  \bibinfo{volume}{30}\/}, \bibinfo{pages}{1514--1541}.
\bibitem[{Zhang \& Li(2019)}]{zhang2019spike}
\bibinfo{author}{Zhang, W.}, \& \bibinfo{author}{Li, P.}
  (\bibinfo{year}{2019}).
\newblock \bibinfo{title}{Spike-train level backpropagation for training deep
  recurrent spiking neural networks}.
\newblock {\it \bibinfo{journal}{Advances in Neural Information Processing
  Systems}\/},  {\it \bibinfo{volume}{32}\/}, \bibinfo{pages}{7802--7813}.
\bibitem[{Zhang \& Li(2020)}]{zhang2020temporal}
\bibinfo{author}{Zhang, W.}, \& \bibinfo{author}{Li, P.}
  (\bibinfo{year}{2020}).
\newblock \bibinfo{title}{Temporal spike sequence learning via backpropagation
  for deep spiking neural networks}.
\newblock {\it \bibinfo{journal}{Advances in Neural Information Processing
  Systems}\/},  {\it \bibinfo{volume}{33}\/}.
\bibitem[{Zheng et~al.(2021)Zheng, Wu, Deng, Hu \& Li}]{zheng2021going}
\bibinfo{author}{Zheng, H.}, \bibinfo{author}{Wu, Y.}, \bibinfo{author}{Deng,
  L.}, \bibinfo{author}{Hu, Y.}, \& \bibinfo{author}{Li, G.}
  (\bibinfo{year}{2021}).
\newblock \bibinfo{title}{Going deeper with directly-trained larger spiking
  neural networks}.
\newblock In {\it \bibinfo{booktitle}{Proceedings of the AAAI Conference on
  Artificial Intelligence}\/} (pp. \bibinfo{pages}{11062--11070}).
\newblock volume~\bibinfo{volume}{35}.

\end{thebibliography}

\newpage
\appendix
\section{Details of Experimental Setup and Hyperparameters}
\label{app:hyper-parameters}
The following techniques were employed in \S~\ref{sec:experiment}.
We used the random augmentation~\citep{cubuk2020randaugment} and dropout~\citep{srivastava2014dropout} for the output layer.
We also adopted the test time augmentation~\citep{krizhevsky2012imagenet} to improve the robustness during inference.

The detailed architectures of employed networks are shown in Table~\ref{tbl:architecture}.
Employed formulas for each method in Table~\ref{tbl:comparison-cifar10-a} are shown in Table~\ref{tbl:eqs-forward-backward}.
As our method in Table~\ref{tbl:eqs-forward-backward}, $u_{\rm th}$ and $\tau_s$ are chosen so that the value of the surrogate gradient at $u = u_{\rm th}$ becomes the same as that for STE~\citep{bengio2013estimating}, and $\alpha$ is the same as~\citet{zhang2020temporal}.
\Add{As for SiBNN, the hyperparameter is set to $\rho=0.3$, and the trainable parameters are initialized by $\theta=0.3, \Delta=1.0$. These settings are the same as in the original paper. As for SLAYER, the hyperparameters are set to $\alpha_{\rm {slayer}}=1.0, \beta_{\rm slayer}=1.5$. Note that the thresholds are the same except STE-b (for BNNs) and ReLU (for ANNs).}
The other hyperparameters were as follows: epochs $= 200$, dropout rate $= 0.1$, and label smoothing $\epsilon = 0.1$.
For ResNet18, a grid search was performed to obtain the optimum result from the following hyperparameters,
\begin{align*}
&\text{optimizer: \{SGD, Adam\}}, \\
&\text{learning rate: \{0.0005, 0.001, 0.005, 0.01, 0.05\}}, \\
&\text{weight decay: \{0.0005, 0.001, 0.005\}},
\end{align*}
where the learning rate was scheduled by the cosine annealing \citep{loshchilov2016sgdr}.
For ResNet106, the same hyperparameters as ResNet18 were used.
{}

\begin{table}[t]
\centering
\caption{Architectures for ResNets: Numbers in the table denote the channel size and the residual blocks. For example, A64 $\times$ 2 indicates the stack of two residual modules of type A (Fig.~\ref{fig:architecture}) whose channel size is 64. Detailed process of each layer is shown in \S~\ref{sec:single-step-neural-networks} and Fig.~\ref{fig:architecture}.}
\begin{tabular}{lcc}
\hline
{}                                & ResNet18        & ResNet106        \\ \hline
Input layer                       & 64 $\times$ 1   & 64    $\times 1$   \\ \hline
Residual block 1                  & A64 $\times$ 2  & A64 $\times$ 13  \\ \hline
\multirow{2}{*}{Residual block 2} & B128 $\times$ 1 & B128 $\times$ 1  \\
                                  & A128 $\times$ 1 & A128 $\times$ 12 \\ \hline
\multirow{2}{*}{Residual block 3} & B256 $\times$ 1 & B256 $\times$ 1  \\
                                  & A256 $\times$ 1 & A256 $\times$ 12 \\ \hline
\multirow{2}{*}{Residual block 4} & B512 $\times$ 1 & B512 $\times$ 1  \\
                                  & A512 $\times$ 1 & A512 $\times$ 12 \\ \hline
Output layer                      & 10 $\times 1$   & 10 $\times 1$  \\ \hline
\end{tabular}
\label{tbl:architecture}
\end{table}

\begin{table}[t]
\caption{Forward and backward formulas for each method in Table~\ref{tbl:comparison-cifar10-a}}
\label{tbl:eqs-forward-backward}
\centering
\begin{tabular}{ccc}
\hline
Method & Forward & Backward \\ \hline
ReLU &
$
a(u)
= \left\{
\begin{array}{ll}
u & (u \geq 0) \\
0 & (u < 0)
\end{array}
\right.
$
&
$
\frac{\partial a}{\partial u}\left(u\right)
= \left\{
\begin{array}{ll}
1 & (u \geq 0) \\
0 & (u < 0)
\end{array}
\right.
$
\\
STE-b &
$
a(u)
= \left\{
\begin{array}{ll}
1 & (u \geq 0) \\
0 & (u < 0)
\end{array}
\right.
$
&
$
\frac{\partial a}{\partial u}\left(u\right)
= \left\{
\begin{array}{ll}
1 & (|u| \leq 1) \\
0 & (|u| > 1)
\end{array}
\right.
$
\\
STE-s &
$
a(u)
= \left\{
\begin{array}{ll}
1 & (u \geq 1) \\
0 & (u < 1)
\end{array}
\right.
$
&
$
\frac{\partial a}{\partial u}\left(u\right)
= \left\{
\begin{array}{ll}
1 & (\Add{|u-1| \leq 1}) \\
0 & (\Add{|u-1| > 1})
\end{array}
\right.
$
\\
SiBNN &
$
a(u)
= \left\{
\begin{array}{ll}
1 & (u \geq \Add{\theta}) \\
0 & (u < \Add{\theta})
\end{array}
\right.
$
&
$
\frac{\partial a}{\partial u}\left(u\right)
= \left\{
\begin{array}{ll}
1 & (\Add{\theta - \rho \Delta} \leq u \leq \Add{\theta + \Delta}) \\
0 & (\text{otherwise})
\end{array}
\right.
$
\\
EENC &
$
a(u)
= \left\{
\begin{array}{ll}
1 & (u \geq 1) \\
0 & (u < 1)
\end{array}
\right.
$
&
$
\frac{\partial a}{\partial u}\left(u\right)
= \max{(1-|u-1|, 0)}
$
\\
SLAYER &
$
a(u)
= \left\{
\begin{array}{ll}
1 & (u \geq 1) \\
0 & (u < 1)
\end{array}
\right.
$
&
$
\frac{\partial a}{\partial u}\left(u\right)
= \Add{\alpha_{\rm {slayer}}} \exp{(-\Add{\beta_{\rm slayer}}|u-1|)}
$
\\
Ours &
$
a(u)
= \left\{
\begin{array}{ll}
1 & (u \geq 1) \\
0 & (u < 1)
\end{array}
\right.
$
&
Eq.~\ref{eqn:ssn-surrogate-gradient}
($u_{\rm th} = 1.0, \tau_s=1.0, \alpha=0.2$)
\\
\hline
\end{tabular}
\end{table}

\begin{figure}[t]
\centering
\begin{minipage}[b]{0.475\linewidth}
    \includegraphics[width=\linewidth]{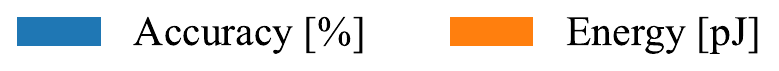}
\end{minipage}
\\
\begin{minipage}[b]{0.475\linewidth}
    \centering
    \includegraphics[width=\linewidth]{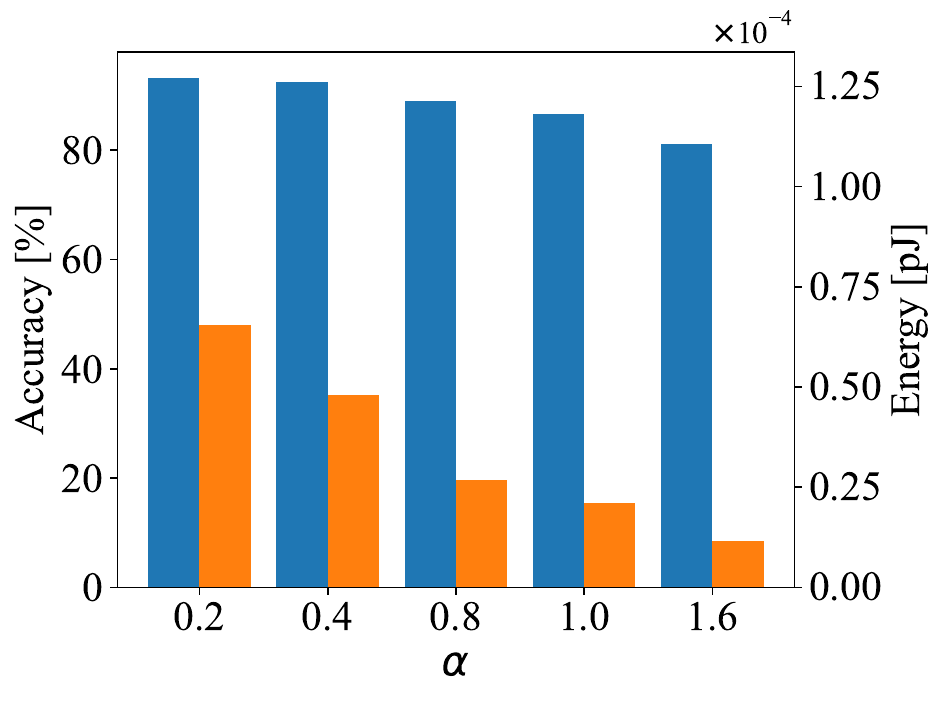}
    \subcaption{S3NN. For fixed $\tau_s=1.0$.}
    \label{fig:comparison-width-fixed-tau}
\end{minipage}
\hfill
\begin{minipage}[b]{0.475\linewidth}
    \centering
    \includegraphics[width=\linewidth]{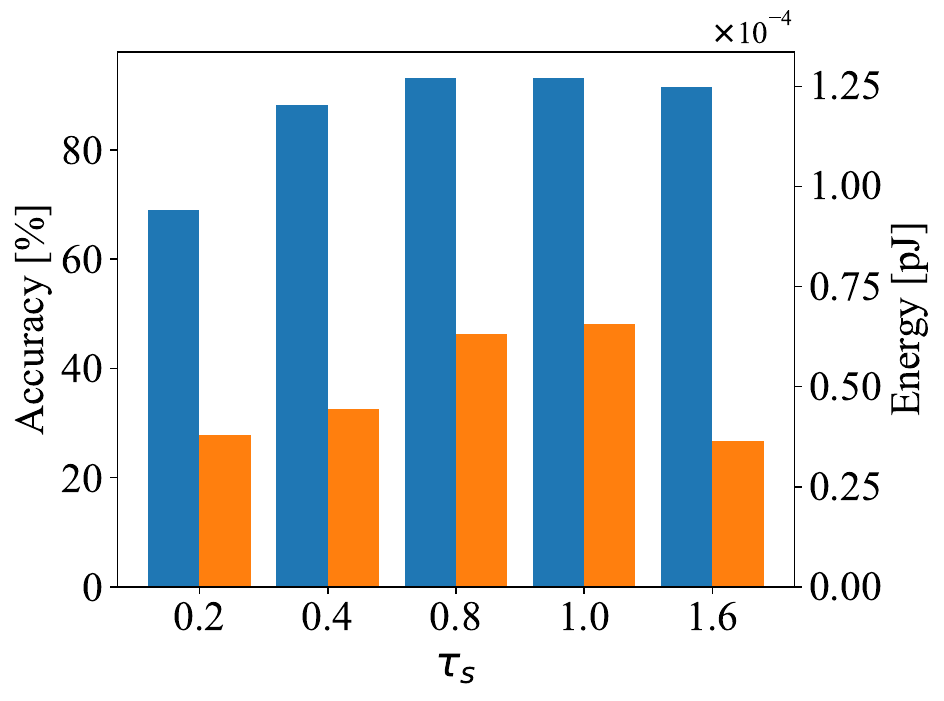}
    \subcaption{S3NN. For fixed $\alpha=0.2$.}
    \label{fig:comparison-width-fixed-alpha}
\end{minipage}
\\
\begin{minipage}[b]{0.475\linewidth}
    \centering
    \includegraphics[width=\linewidth]{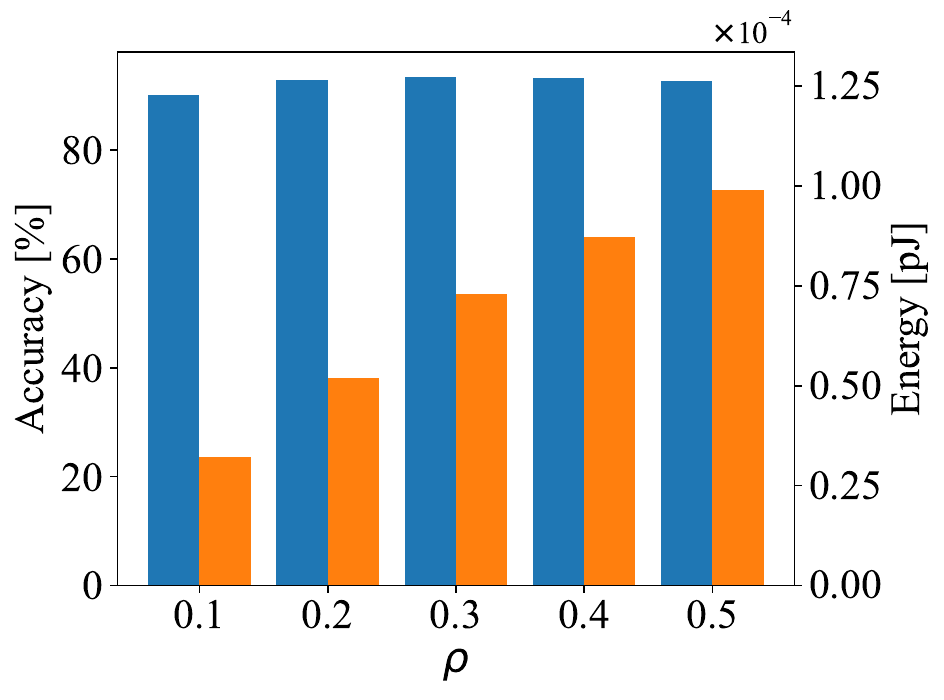}
    \subcaption{SiBNN.}
    \label{fig:comparison-width-sibnn}
\end{minipage}
\hfill
\begin{minipage}[b]{0.475\linewidth}
    \centering
    \includegraphics[width=\linewidth]{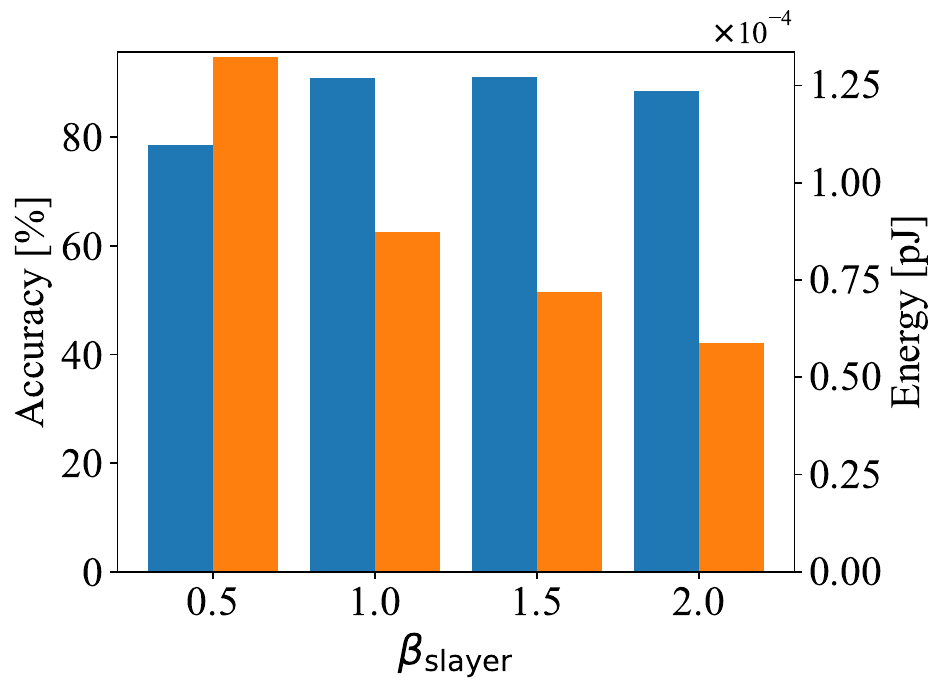}
    \subcaption{SLAYER. For fixed $\alpha_{\rm slayer}=1.0$.}
    \label{fig:comparison-width-slayer}
\end{minipage}
\caption{\Add{Influence of hyperparameters on the accuracy and energy consumption
(CIFAR-10, ResNet18)}}
\label{fig:comparison-width}
\end{figure}

\Erase{Fig.~\ref{fig:comparison-width} shows}\Add{Figs.~\ref{fig:comparison-width-fixed-tau} and \ref{fig:comparison-width-fixed-alpha} show the} influence of the hyperparameters $\alpha$ and $\tau_s$ \Erase{on the accuracy and energy consumption for the CIFAR-10 dataset. }\Add{related to the width of our surrogate gradient.} These figures show that accuracy and energy consumption vary depending on \Erase{these hyperparameters}\Add{them}.
\Add{For example, accuracy and energy decrease when $\alpha$ increases. It implies that these hyperparameters control the trade-off between accuracy and energy efficiency, {\it{i.e.}}, they cause high energy consumption to keep high accuracy, and vice versa.} 
Finding the optimal hyperparameters is included in our future work.
\Add{\Add{Figs.~\ref{fig:comparison-width-sibnn} and \ref{fig:comparison-width-slayer} also show the influence of the hyperparameters of SiBNN and SLAYER related to the width of the surrogate gradients. They also control the trade-off between accuracy and energy efficiency. Note that our choice of values for $\rho$ and $\beta_{\rm slayer}$ in Table~\ref{tbl:comparison-cifar10-a} correspond to the maximum accuracy.}}


\section{Details of Energy Calculation}
\label{app:energy-calculation}
The actual values for energy consumption are out of the scope of this study.
However, we can estimate the energy consumption as follows.

\begin{figure}[t]
    \centering
    \includegraphics[width=120mm]{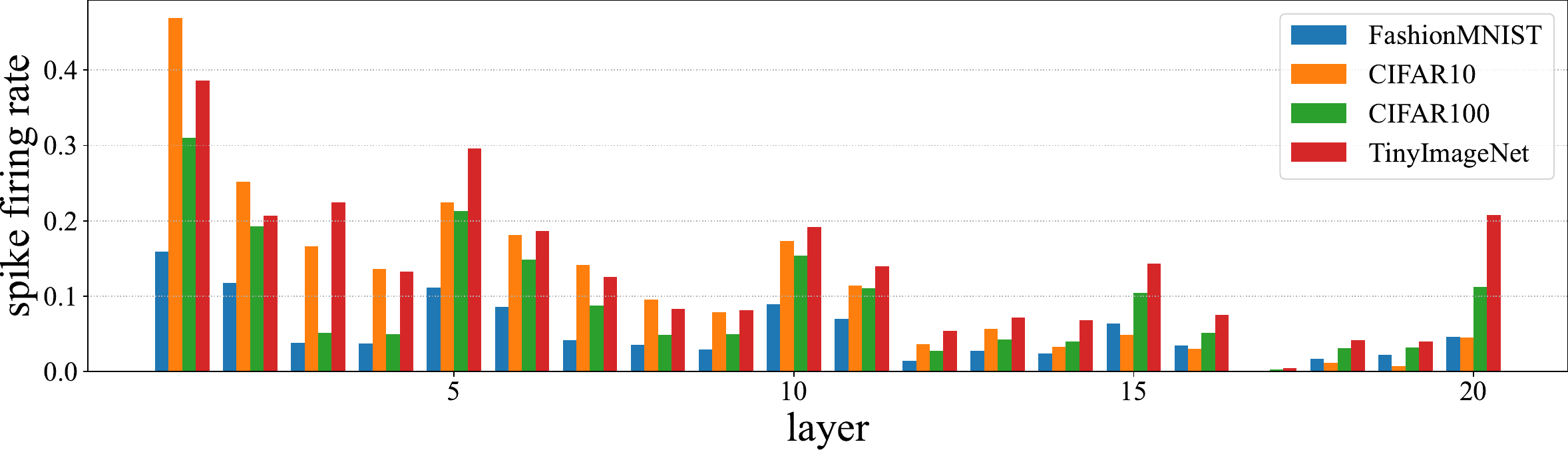}
\caption{Example of $R_{\text {Method}}(l)$: Each bar represents $R_{\text {Method}}(l)$ of the proposed method in ResNet18 for each dataset.}
\label{fig:spike-rate}
\end{figure}

First, we define ${\text {FLOPs}}(l)$, the number of floating-point operations in $l$-th layer as follows:
\begin{align}
{\text {FLOPs}}(l)
= \left\{
\begin{array}{ll}
{k^{(l)}}^2 \times W^{(l)} \times H^{(l)} \times C_{\text {in}}^{(l)} \times C_{\text {out}}^{(l)} & ({\text {for Convolutional layer}}) \\
C_{\text {in}}^{(l)} \times C_{\text {out}}^{(l)} & ({\text {for Linear layer}}).
\end{array}
\right.
\end{align}
Here, $C_{\text {in}}^{(l)}$ and $C_{\text {out}}^{(l)}$ are the input and output channel sizes, respectively, $k^{(l)}$ represents the kernel size, and $W^{(l)}$ and $H^{(l)}$ are the width and height of the feature map.
We can compute the energy consumption estimations for the full-precision network or each model by multiplying ${\text {FLOPs}}(l)$ and $E_{\text {MAC}}$ or $E_{\text {AC}}$, where $E_{\text {MAC}}$ or $E_{\text {AC}}$ is the energy consumption per MAC operation or additive operation.
For $E_{\text {MAC}} = 4.6$ [pJ], $E_{\text {AC}} = 0.9$ [pJ] (45 nm CMOS processor~\citep{horowitz20141}), the energy efficiency by binarizing the activation function is $E_{\text {MAC}}/E_{\text {AC}} \sim 5.11$.
Furthermore, to estimate the energy efficiency using asynchronous processing assumed in the SNNs field, we can compute the spike firing rate $R_{\text {Method}}(l)$ for each layer:
\begin{align}
R_{\text {Method}}(l)
= \mathbb{E} \left[
\frac{\text {\# spikes of {\it {l}}-th layer}}{\text {\# neurons of {\it {l}}-th layer}} \right].
\end{align}
Here, spikes represent neurons such that $u_{i}^{(l)} \geq u_{\rm th}$.
Fig.~\ref{fig:spike-rate} shows an example of $R_{\text {Method}}(l)$ of the proposed method.
Using them, we can compute the energy consumption for the full-precision network and each method as follows:
\begin{align}
E_{\text {Full}} &= E_{\text {MAC}} \sum_l {\text {FLOPs}}(l), \label{eqn:energy-full}\\
E_{\text {Method}} &= E_{\text {AC}} \sum_l {\text {FLOPs}}(l) \times R_{\text {Method}}(l).
\end{align}
These are the energies it takes to obtain the output given a single image. Also, when we use models with the time step $T > 1$, the corresponding energies are as follows.
\begin{align}
E_{\text {Full}} &= T \times E_{\text {MAC}} \sum_l {\text {FLOPs}}(l), \label{eqn:energy-full}\\
E_{\text {Method}} &= T \times E_{\text {AC}} \sum_l {\text {FLOPs}}(l) \times R_{\text {Method}}(l).
\end{align}

\section{Classification benchmarks}
\label{app:classification-benchmarks}

Tables~\ref{tbl:comparison-fmnist-s}--\ref{tbl:comparison-tinyimagenet-s} compare the proposed method and related BNN and SNN methods for each dataset. Note that the BSNN in ``Paradigm'' means an SNN whose weights are binarized, and the NAS in ``Architecture" means a neural architecture searched model. As shown in these tables, we can observe similar trends as in Table~\ref{tbl:comparison-cifar10-s}.

\begin{table}[t]
\caption{Classification benchmarks for the Fashion-MNIST dataset. The description is the same as Table~\ref{tbl:comparison-cifar10-s}.}
\label{tbl:comparison-fmnist-s}
\centering
\begin{tabular}{lllrl} \hline
{} & Paradigm & Architecture & Time Steps & Accuracy [\%] \\ \hline
\cite{qiao2020stbnn} & BNN & 3Linears & 1 & 87.0 \\
\cite{kheradpisheh2020bs4nn} & BSNN & 1Linear & 256 & 87.3 \\
\cite{zhang2019spike} & SNN & 1Linear.1Recurrent & 400 & 90.13 \\
\cite{cheng2020lisnn} & SNN & 2Convs.1Linear & 20 & 92.07 \\
\cite{zhang2020temporal} & SNN & 2Convs.1Linear & 5 & 92.83 (91.6) \\
\cite{fang2020incorporating} & SNN & 2Convs.2Linears & 8 & 94.38 \\
{\bf {This work}} & S$^3$NN & ResNet106 & 1 & {\bf {94.5 (94.06)}} \\
{\bf {This work}} & S$^3$NN & ResNet18 & 1 & {\bf {94.81 (95.03)}} \\
\hline
\end{tabular}
\end{table}
\begin{table}[t]
\caption{Classification benchmarks for the CIFAR-100 dataset. The description is the same as Table~\ref{tbl:comparison-cifar10-s}.}
\label{tbl:comparison-cifar100-s}
\centering
\begin{tabular}{lllrl} \hline
{} & Paradigm & Architecture & Time Steps & Accuracy [\%] \\ \hline
\cite{chen2021bnn} & BNN & ReActNet18 & 1 & 68.34 (68.79) \\
\cite{shen2020balanced} & BNN & ResNet20 & 1 & 69.38 \\
\cite{lu2020exploring} & BSNN & modified VGG15 & 62 & 62.07 (64.9) \\
\cite{esser2016cover} & SNN & 3Convs.8Linears & 1 & 55.64 \\
\cite{kim2022privatesnn} & SNN & VGG16 & 200 & 62.3 (64.3) \\
\cite{kundu2021towards} & SNN & VGG11 & 10 & 65.34 (67.4) \\
\cite{rathi2019enabling} & SNN & VGG11 & 125 & 67.87 (71.21) \\
\cite{na2022autosnn} & SNN & (NAS) & 8 & 69.16 \\
\cite{chowdhury2021one} & SNN & VGG16 & 1 & 70.15 (72.46) \\
\cite{deng2020optimal} & SNN & VGG16 & 128 & 70.47 \\
\cite{han2020rmp} & SNN & VGG16 & 2048 & 70.93 (71.22) \\
{\bf {This work}} & S$^3$NN & ResNet18 & 1 & {\bf {71.25 (71.78)}} \\
\cite{yan2021near} & SNN & VGG* & 300 & 71.84 (71.84) \\
\cite{kim2022neural} & SNN & (NAS) & 5 & 73.04 \\
{\bf {This work}} & S$^3$NN & ResNet106 & 1 & {\bf {73.56 (70.04)}} \\
\cite{pmlr-v139-li21d} & SNN & ResNet20 & 32 & 76.32 (77.16) \\
\cite{li2021bsnn} & SNN & ResNet20 & 265 & 78.12 (77.97) \\
\hline
\end{tabular}
\end{table}
\begin{table}[t]
\caption{Classification benchmarks for the TinyImageNet dataset. The description is the same as Table~\ref{tbl:comparison-cifar10-s}.}
\label{tbl:comparison-tinyimagenet-s}
\centering
\begin{tabular}{lllrl} \hline
{} & Paradigm & Architecture & Time Steps & Accuracy [\%] \\ \hline
\cite{na2022autosnn} & SNN & (NAS) & 8 & 46.79 \\
\cite{kim2022privatesnn} & SNN & VGG16 & 200 & 51.9 (50.7) \\
\cite{kundu2021towards} & SNN & VGG16 & 10 & 54.1 (57.0) \\
\cite{kim2022neural} & SNN & (NAS) & 5 & 54.6 \\
{\bf {This work}} & S$^3$NN & ResNet18 & 1 & {\bf {55.49 (58.05)}} \\
\cite{kim2020revisiting} & SNN & VGG11 & 30 & 57.8 \\
\hline
\end{tabular}
\end{table}

\end{document}